\documentclass[10pt,twocolumn,letterpaper]{article}

\usepackage{cvpr}            %

\usepackage[dvipsnames]{xcolor}

\usepackage{amsmath}
\usepackage{placeins}

\makeatletter%
\providecommand\color[2][]{%
  \errmessage{(Inkscape) Color is used for the text in Inkscape, but the package 'color.sty' is not loaded}%
  \renewcommand\color[2][]{}%
}%
\providecommand\transparent[1]{%
  \errmessage{(Inkscape) Transparency is used (non-zero) for the text in Inkscape, but the package 'transparent.sty' is not loaded}%
  \renewcommand\transparent[1]{}%
}%
\newcommand*\fsize{\dimexpr\f@size pt\relax}%
\newcommand*\lineheight[1]{\fontsize{\fsize}{#1\fsize}\selectfont}%
\ifx\svgwidth\undefined%
  \ifx\svgscale\undefined%
    \relax%
  \else%
  \fi%
\else%
\fi%
\global\let\svgwidth\undefined%
\global\let\svgscale\undefined%
\makeatother%

\definecolor{cvprblue}{rgb}{0.21,0.49,0.74}
\usepackage[pagebackref,breaklinks,colorlinks,citecolor=cvprblue]{hyperref}

\def\paperID{396} %
\def\confName{3DV\xspace}
\def\confYear{2024\xspace}

\newcommand{\suppTitle}[1]{\renewcommand{\suppTitle}{#1}}
\suppTitle{ Visual Tomography:\\ Physically Faithful Volumetric Models of Partially Translucent Objects}

\title{Visual Tomography:\\Physically Faithful Volumetric Models of Partially Translucent Objects}

\author{$^{*\dagger}$David Nakath,  $^\dagger$Xiangyu Weng, $^{*\dagger}$Mengkun She, $^{*\dagger}$Kevin K\"oser\vspace{.2cm}\\
\begin{tabular}{cc}
    $^{*}$Marine Data Science & $^{\dagger}$Oceanic Machine Vision\\
    University of Kiel & GEOMAR -- {\small Helmholtz Centre for Ocean Research Kiel}\\
    Christian-Albrechts-Platz 4, 24118 Kiel, Germany & Wischhofstr. 1-3, 24148 Kiel, Germany
\end{tabular}~\\
{\tt\small dna@informatik.uni-kiel.de}~\\
{\hspace{-.35cm}
\normalfont \small
\def\svgwidth{1\linewidth}
\begin{picture}(1,0.23145894)%
  \lineheight{1}%
  \setlength\tabcolsep{0pt}%
  \put(0,0){\includegraphics[width=\unitlength,page=1]{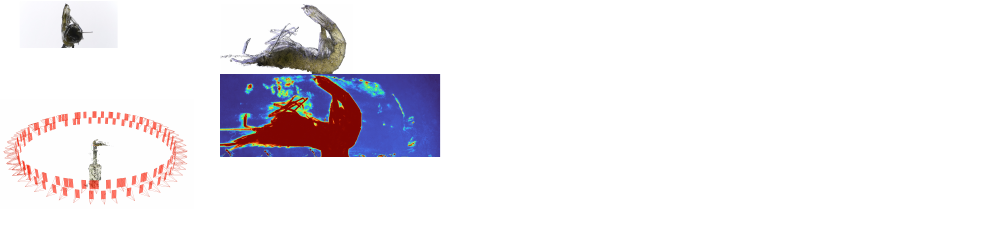}}%
  \put(0.00233556,0.16648532){\color[rgb]{0,0,0}\makebox(0,0)[lt]{\lineheight{1.25}\smash{\begin{tabular}[t]{l}Brightfield Images\end{tabular}}}}%
  \put(0,0){\includegraphics[width=\unitlength,page=2]{pipeline.pdf}}%
  \put(0.08191126,0.12933787){\color[rgb]{0,0,0}\makebox(0,0)[lt]{\lineheight{1.25}\smash{\begin{tabular}[t]{l}Intrinsics\end{tabular}}}}%
  \put(0.07812235,0.00376361){\color[rgb]{0,0,0}\makebox(0,0)[lt]{\lineheight{1.25}\smash{\begin{tabular}[t]{l}Extrinsics\end{tabular}}}}%
  \put(0,0){\includegraphics[width=\unitlength,page=3]{pipeline.pdf}}%
  \put(1.31684131,0.00128299){\color[rgb]{0,0,0}\makebox(0,0)[lt]{\lineheight{1.25}\smash{\begin{tabular}[t]{l}Slicing\end{tabular}}}}%
  \put(0,0){\includegraphics[width=\unitlength,page=4]{pipeline.pdf}}%
  \put(0.003913,0.00438655){\color[rgb]{0,0,0}\makebox(0,0)[lt]{\lineheight{1.25}\smash{\begin{tabular}[t]{l}SfM\end{tabular}}}}%
  \put(0,0){\includegraphics[width=\unitlength,page=5]{pipeline.pdf}}%
  \put(0.47723721,0.00351995){\color[rgb]{0,0,0}\makebox(0,0)[lt]{\lineheight{1.25}\smash{\begin{tabular}[t]{l}Differentiable Raytracing\end{tabular}}}}%
  \put(0,0){\includegraphics[width=\unitlength,page=6]{pipeline.pdf}}%
  \put(0.47819795,0.20392499){\color[rgb]{0,0,0}\makebox(0,0)[lt]{\lineheight{1.25}\smash{\begin{tabular}[t]{l}Phase\end{tabular}}}}%
  \put(0.47742581,0.15608854){\color[rgb]{0,0,0}\makebox(0,0)[lt]{\lineheight{1.25}\smash{\begin{tabular}[t]{l}Albedo\end{tabular}}}}%
  \put(0.47634679,0.10706604){\color[rgb]{0,0,0}\makebox(0,0)[lt]{\lineheight{1.25}\smash{\begin{tabular}[t]{l}Density\end{tabular}}}}%
  \put(0.22572189,0.21208714){\color[rgb]{0,0,0}\makebox(0,0)[lt]{\lineheight{1.25}\smash{\begin{tabular}[t]{l}Shape\end{tabular}}}}%
  \put(0,0){\includegraphics[width=\unitlength,page=7]{pipeline.pdf}}%
  \put(0.22640057,0.13744586){\color[rgb]{1,1,1}\makebox(0,0)[lt]{\lineheight{1.25}\smash{\begin{tabular}[t]{l}Density\end{tabular}}}}%
  \put(0.754815,0.20955523){\color[rgb]{1,1,1}\makebox(0,0)[lt]{\lineheight{1.25}\smash{\begin{tabular}[t]{l}Immerse\end{tabular}}}}%
  \put(0,0){\includegraphics[width=\unitlength,page=8]{pipeline.pdf}}%
  \put(0.75627556,0.0954497){\color[rgb]{1,1,1}\makebox(0,0)[lt]{\lineheight{1.25}\smash{\begin{tabular}[t]{l}Darkfield\end{tabular}}}}%
  \put(0,0){\includegraphics[width=\unitlength,page=9]{pipeline.pdf}}%
  \put(0.22592164,0.00350527){\color[rgb]{0,0,0}\makebox(0,0)[lt]{\lineheight{1.25}\smash{\begin{tabular}[t]{l}NeRF\end{tabular}}}}%
  \put(0,0){\includegraphics[width=\unitlength,page=10]{pipeline.pdf}}%
  \put(0.90246237,0.10962724){\color[rgb]{1,1,1}\makebox(0,0)[lt]{\lineheight{1.25}\smash{\begin{tabular}[t]{l}Re-Render\end{tabular}}}}%
  \put(0,0){\includegraphics[width=\unitlength,page=11]{pipeline.pdf}}%
\end{picture}%
}\vspace{-.5cm}
}

\begin{document}
\maketitle
\sloppy
\begin{abstract}
	When created faithfully from real-world data, Digital 3D representations of objects can be useful for human or computer-assisted analysis. Such models can also serve for generating training data for machine learning approaches in settings where data is difficult to obtain or where too few training data exists, e.g. by providing novel views or images in varying conditions. While the vast amount of visual 3D reconstruction approaches focus on non-physical models, textured object surfaces or shapes, in this contribution we propose a volumetric reconstruction approach that obtains a physical model including the interior of partially translucent objects such as plankton or insects. Our technique photographs the object under different poses in front of a bright white light source and computes absorption and scattering per voxel. It can be interpreted as visual tomography that we solve by inverse raytracing. We additionally suggest a method to convert non-physical NeRF media into a physically-based volumetric grid for initialization and illustrate the usefulness of the approach using two real-world plankton validation sets, the lab-scanned models being finally also relighted and virtually submerged in a scenario with augmented medium and illumination conditions.
	Please visit the project homepage at \url{www.marine.informatik.uni-kiel.de/go/vito}
\end{abstract}

\section{Introduction}
\label{sec:intro}
\begin{figure*}[t]
	\begin{center}
		\includegraphics[width=.24\linewidth]{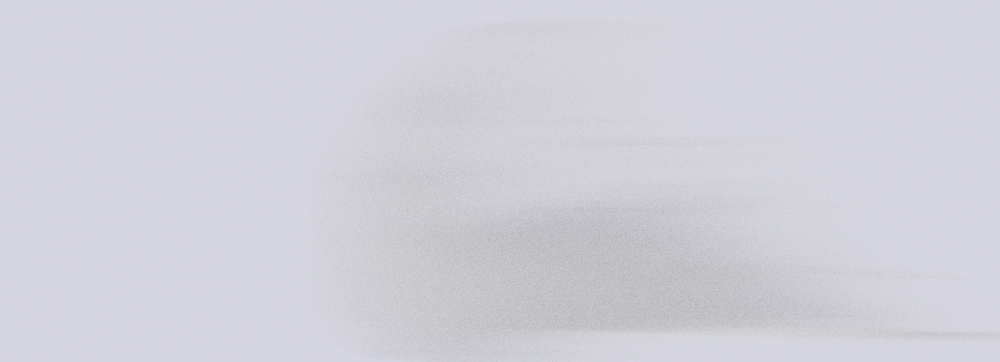}
		\includegraphics[width=.24\linewidth]{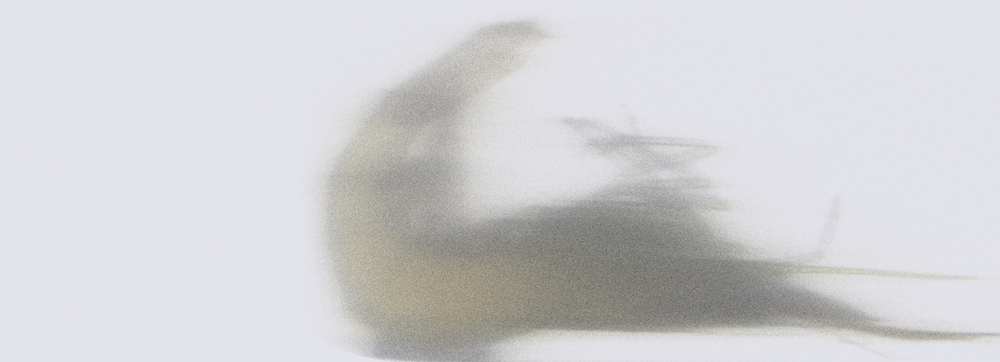}
		\includegraphics[width=.24\linewidth]{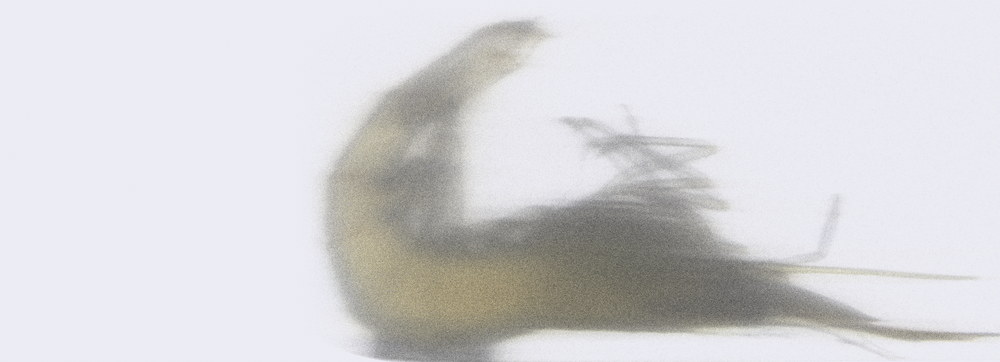}
		\includegraphics[width=.24\linewidth]{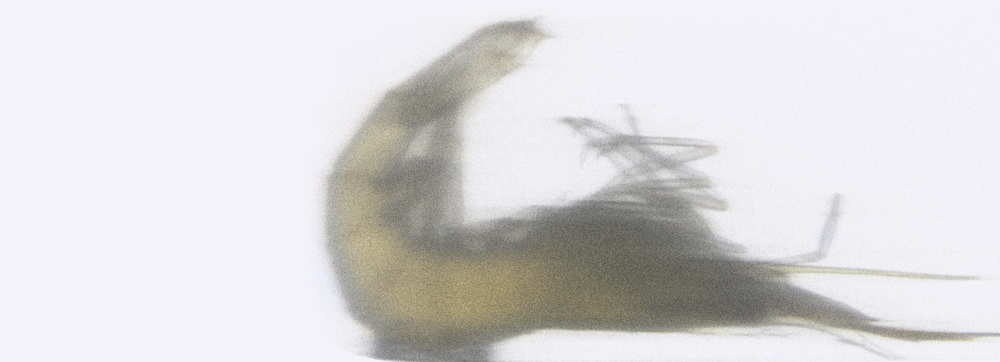}
		\includegraphics[width=.24\linewidth]{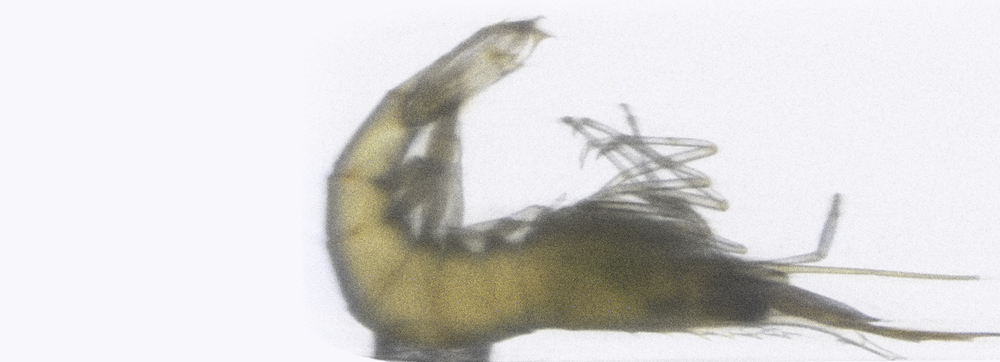}
		\includegraphics[width=.24\linewidth]{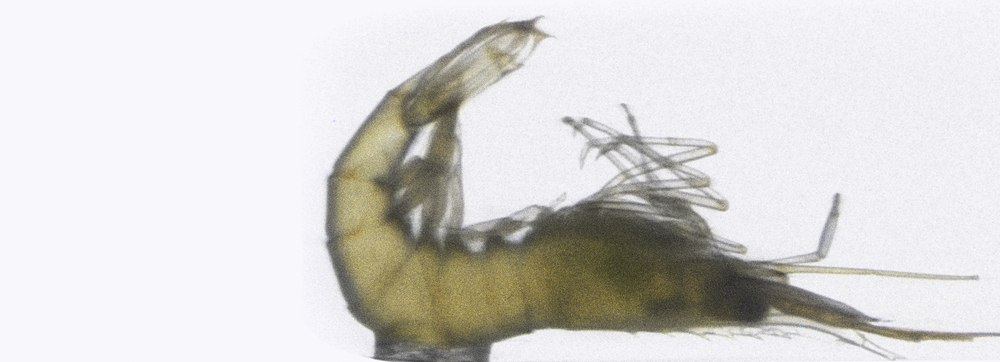}
		\includegraphics[width=.24\linewidth]{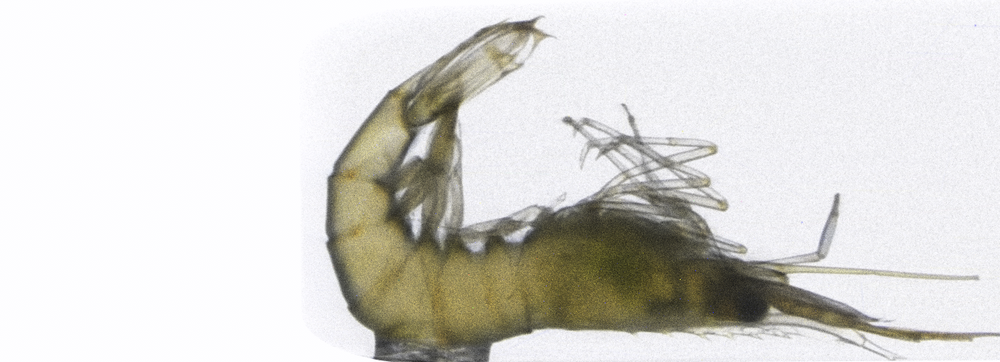}
		\includegraphics[width=.24\linewidth]{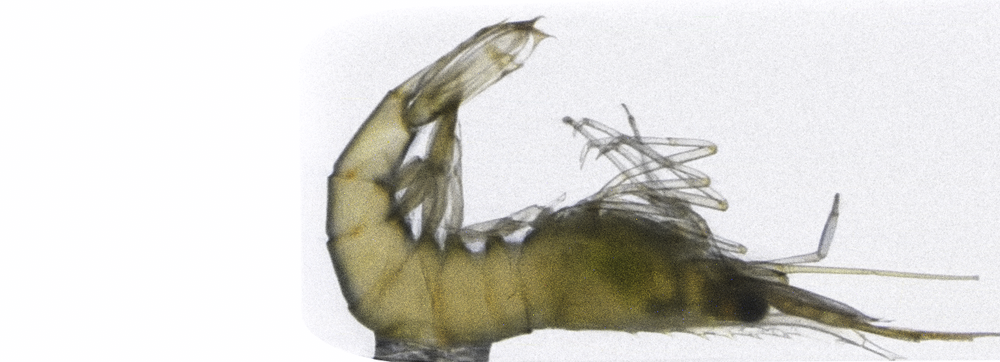}
	\end{center}
	\caption{Dwarf Prawns \textbf{MO + LO} optimization over time, from up left to right down: iteration 1, 2, 3, 4, 10, 20, 40, 60.}
	\label{fig:dp_opt}
\end{figure*}

\begin{figure*}[t]
	\begin{center}
		\includegraphics[width=.24\linewidth]{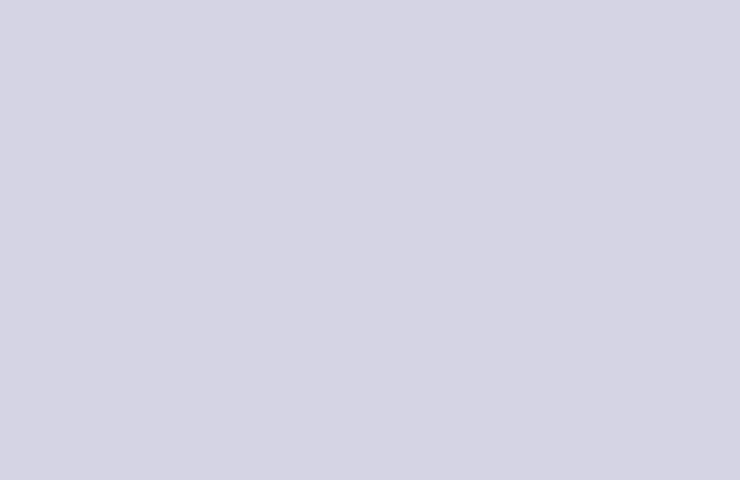}
		\includegraphics[width=.24\linewidth]{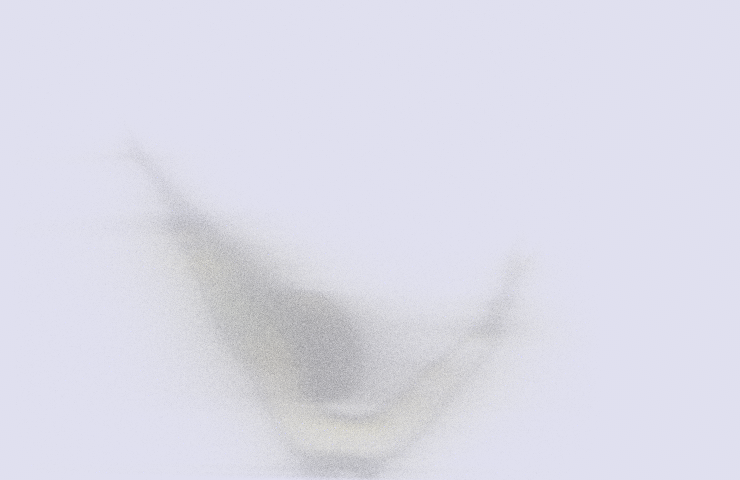}
		\includegraphics[width=.24\linewidth]{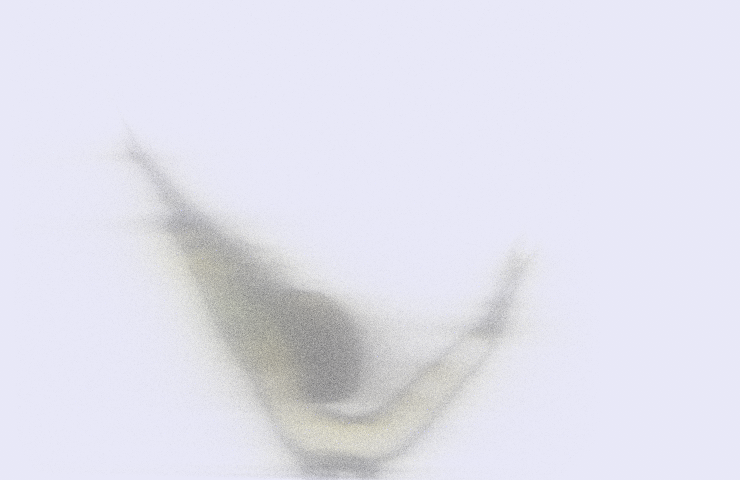}
		\includegraphics[width=.24\linewidth]{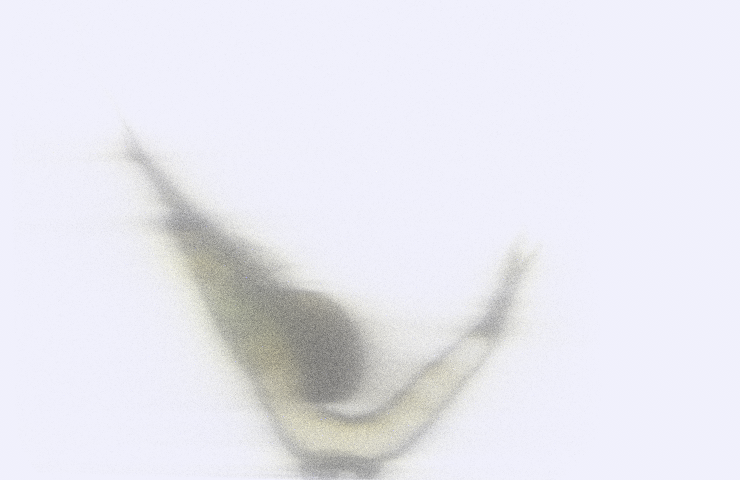}
		\includegraphics[width=.24\linewidth]{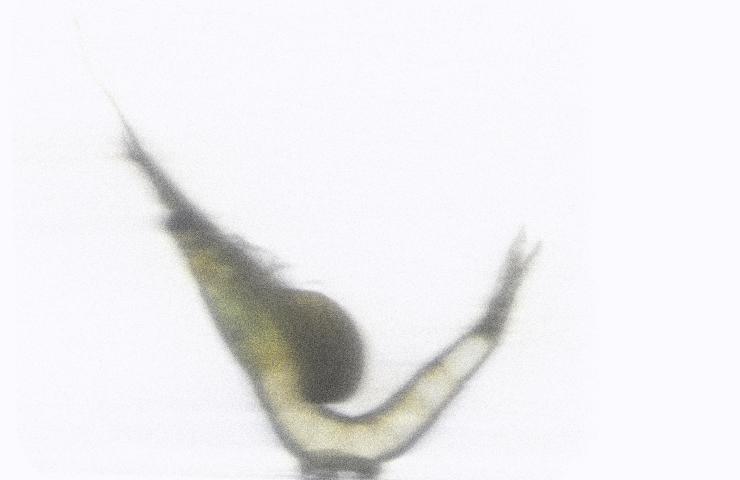}
		\includegraphics[width=.24\linewidth]{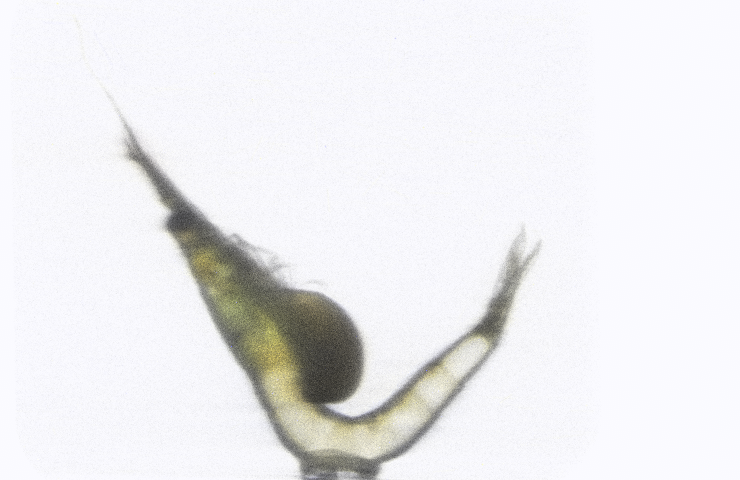}
		\includegraphics[width=.24\linewidth]{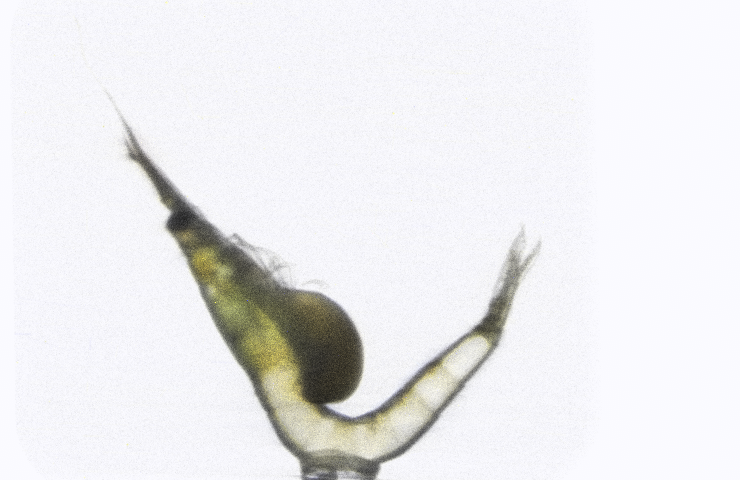}
		\includegraphics[width=.24\linewidth]{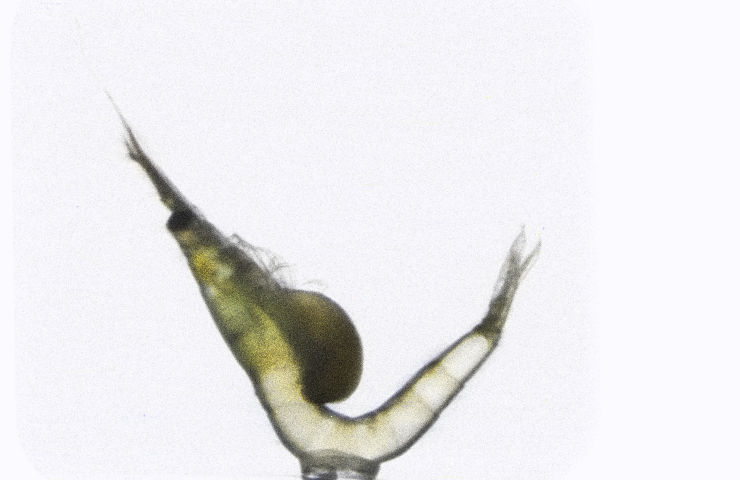}
	\end{center}
	\caption{Praunous Flexuosus \textbf{MO + LO} optimization over time, from up left to right down: iteration 1, 2, 3, 4, 10, 20, 40, 60.}\label{fig:pf_opt}
\end{figure*}

Obtaining faithful 3D models from images is a key technology that is required by many applications in science and engineering.
While approaches based on visual light have largely focused on 3D surface models, e.g. \cite{kutulakos99spacecarving,newcombe2011kinectfusion, schoenberger2016sfm,schoenberger2016mvs}, much less attention has been paid to the interior of (partially) translucent objects.
Such objects are very common in the micro world and include life in the ocean, insects or small organisms, various kinds of tissue and samples in biology, geology or medicine, but also larger things like gems (amber), certain kinds of food, household items, industrial materials and even model organisms like zebrafish that are studied in drug development.
Our key motivation for this work is to obtain photo-realistic, volumetric 3D models of oceanic zooplankton like copepods%
, such that their (semi-transparent) appearance can be simulated for a variety of environmental and observation conditions. Such synthetic visual models of ocean organisms are expected to improve in-situ classification or compression algorithms for low-bandwidth transmission, in particular for species underrepresented in visual training data.
Monitoring the abundance of key species (\emph{essential ocean variables}) can serve as a proxy to understanding the health state of the ocean, but obtaining good training data for machine learning \cite{irisson2022mlplankton} is tedious and expensive and notoriously incomplete, as optical appearance can drastically vary with different waters, depths and illumination scenarios \cite{nakath2022optical}.
Our goal is therefore to scan key species in the lab, later virtually submerge the obtained 3D models in various kinds of optically well-understood ocean waters  \cite{preisendorfer1964physical,jerlov}, and finally use virtual illumination and observation conditions of a particular ocean sensor (such as strobe light, laser sheet illumination \cite{picheral21uvp6} or a ring of LEDs \cite{li22buoy}).

While recently, neural radiance fields (NeRFs \cite{mildenhall2021nerf}) have become popular to learn the appearance of 3D scenes and objects, such representations are not directly suitable for replacing the medium and adding artificial light sources. We show in this paper that they can however be used to initialize explicit, physically meaningful, volumetric models that are used in volumetric raytracing.

The appearance of our objects is largely characterized by the amount of light that is absorbed and scattered \emph{by internal volumetric structure} of the object\footnote{We do not consider refractive effects in this contribution.}.
Our approach therefore relates to computed tomography (CT) \cite{buzug2008computerdtomography,ruckert2022neat} approaches in medicine that use expensive devices to emit and record x-ray radiation in order to obtain information of the interior of the human body. In contrast, we present the object in front of a bright light source and record how much light passes through the object using an off-the-shelf camera.
This is why we refer to our technique as \emph{Visual Tomography} (tomos: "slice").
In this paper, we make the following novel contributions:
\begin{itemize}
	\item A hardware setup and scanning procedure to obtain pose-referenced bright-field imagery of plankton-like translucent objects using just a standard DSLR camera.
	\item A complete method based on differentiable raytracing for physically faithful volumetric reconstruction from real images, including calibration, image acquisition, pose estimation, volume and light optimization and re-rendering of tiny, partially translucent objects.
	\item A NeRF-based initialization step, based on a mapping between a non-physical emissive volume to a physically-based volume-representation.
	\item And several options to improve and augment the data acquired in the optimization steps.
\end{itemize}
To the best of our knowledge this is the first visual tomography approach and the first system to reconstruct the interior of plankton-like translucent objects.

\section{Related Work}
\subsection{Plankton Scanning Devices}
Tasks like observing and scanning of plankton in-situ are challenging due to the translucent nature of plankton.
Current state-of-the-art in-situ plankton abundance measurement devices are the UVP series \cite{picheral2010underwater}.
They use a 638nm laser light source to illuminate a part of the water body for dark field imaging.
The Planktoscope \cite{pollina2022planktoscope}, which is a modular and portable device, uses brightfield illumination to achieve plankton observation.
ZooScan \cite{gorsky2010digital} is designed for digital imaging of preserved zooplankton, which also uses bright field illumination.
Scale-free vertical tracking microscopy \cite{krishnamurthy2020scale}, uses a high-frequency light source to achieve bright-field illumination for the observation of vertical motions of plankton.
All devices are designed to capture snapshots of the objects, and to the best of our knowledge, no system exists that can establish volumetric 3D models of plankton with off-the-shelf camera equipment.

\subsection{Macro and Tiny Object Reconstruction}
Probably one of the closest works in the literature is the DISC3D project \cite{stroebel2018insects}, where insects are digitized using macro photography and classical structure-from-motion techniques. The approach is similar to the first step in our processing pipeline, where we identify the camera poses relative to the object, but DISC3D does not attempt to reconstruct the interior of objects and aims at opaque insects.

Generally, when photographing tiny objects, camera calibration becomes challenging because of the limited depth of field. It can nevertheless be achieved using the approach of \cite{weng2021macal}, who have also shown that focus stacking converts perspective images into an affine camera model, moving the center of projection to infinity. We avoid this effect by choosing a small aperture, effectively trading depth resolution (depth of field) for lateral resolution (potential diffraction).

\subsection{Computed Tomography}
In medicine, computed tomography is a standard tool nowadays to obtain a 3D representation of the human body, e.g. from x-ray radiation. Here, a radiation source is typically placed opposite to a detector at distance $s$, the human body in between. In the simplest, attenuation-only, model, the intensity $I(s)$ measured at the detector is then
\begin{equation}
	I(s) = I(0) e^{- \int_0^s \sigma(x) dx},
\end{equation}
where $\sigma(x)$ is the attenuation coefficient at position $x$ that is sought. However, the detector observes only the integral quantity $I(s)$, and from one line of sight there are infinitely many $\sigma(x)$ configurations that would explain the measurement.
However, a key challenge is that in a $100 \times 100 \times 100$ volume, one million attenuation parameters have to be estimated, which is computationally challenging.
Radon \cite{radon1917integralwerte} proposed the first direct solution to solve the integrals using an approach that was later named the Radon transform. In a recent textbook, Buzug \cite{buzug2008computerdtomography} discusses pros and cons of the various postprocessing methods and alternatives nowadays in detail. Recently, traditional methods for x-ray tomography have also been challenged by neural methods, e.g. \cite{ruckert2022neat}.

\subsection{Neural Scene Representations for Visual Data}
Recently, there has been a lot of progress in using differentiable rendering applied to analysis-by-synthesis approaches \cite{kato2020differentiable}.
Among these, Neural Volumes \cite{Lombardi_2019} and notably NeRFs (Neural Radiance Fields) \cite{mildenhall2021nerf,barron2022mip,muller2022instant} stand out by employing implicit, coordinate-based neural networks to represent 3D volumes.
By leveraging calibrated 2D images for optimization, they demonstrate remarkable capabilities in achieving photorealistic novel viewpoint rendering, while simultaneously capturing intricate scene geometry and view-dependent effects.
To make the scene representation differentiable, NeRFs relax the classic binary occupied/empty space classification of spatial locations by a continuous density, which together with %
an emitted radiance within the 3D volume, theoretically allows also to represent transparency.
Capitalizing on this capability, Dex-NeRF \cite{IchnowskiAvigal2021DexNeRF} and GraspNeRF \cite{dai2023graspnerf} have demonstrated NeRF's potential for enabling robots to manipulate transparent objects within a scene.
However, it is important to note that the non-physical emissive medium representations employed in NeRFs are inherently not re-lightable.
To address this, Bi et al. and Srinivasan et al. \cite{bi2020neural,srinivasan2021nerv} propose an alternative approach to train a neural reflectance field and Wang et al. learn a neural implicit surface by volume rendering \cite{wang2021neus}.
This shift facilitates re-lightable novel viewpoint synthesis, however, they are targeting textured surfaces instead of transparent volumes.
In \cite{yu2023learning}, re-lightable volumes are represented by learned object-centric neural scattering functions.
However, this approach depends on a distant directional light source, which does not apply to our setup.

Additionally, it is inherently difficult to replace the medium (e.g. turbid waters), that is sort of burnt into NeRF models, after acquisition.
Therefore, we build upon the NeRF as a bootstrapping method in a twofold way:
(i) we use a non-physical emissive volume to bootstrap the density values of a physically-based medium representation. While \cite{nimierdavid2022unbiased} used a simplified NeRF which works directly on the wide-bandwidth densities of a discrete voxel-grid, we present a method to directly initialize such a voxel grid from the scalar density predicted by a classic NeRF MLP  (see Eq. \ref{eq:nerf_mlp}) by a rule as given in Eq. \ref{eq:inv_em}.
(ii) we extract the shape of the specimen and re-use it as a 3D boundary between the medium-based specimen representation and a physically meaningful water-representation.
In this work, we employ Nerfacto \cite{tancik2023nerfstudio}, which incorporates the latest findings from multiple research papers from the NeRF community \cite{barron2022mip,muller2022instant,martin2021nerf}.

\subsection{(Differentiable) Physically-Based Raytracing}
Building on the foundation of geometric optics \cite{preisendorfer1965radiative}, a principled  medium representation can be established within the framework of physically-based raytracing (PBRT) \cite{pharr2023physically}.
The latter significantly matured in the theoretical as well as in the hardware-dimension.
Furthermore, the PBRT model allows for direct re-parametrizations of the scene with actual measured physical parameters.

First attempts to differentiate PBRT approaches used off-the shelf PBRT software and ported them to autodiff frameworks \cite{che2018inverse}.
However, such an approach implemented, in a naive manner, entails severe performance issues and theoretical shortcomings like e.g., with respect to the handling of visibility changes.
The latter were solved by edge sampling in  one of the first PBRT-approaches directly tailored to differentiability \cite{li2018differentiable}.
In addition Mitsuba 2 was published in \cite{nimier2019mitsuba}, which rests on top of the Enoki autodiff library \cite{Enoki}.
The latter particularly strives to tackle the performance issues related to autodiff-gradients of raytracing approaches.

Still, all previously mentioned implementations suffered from severe performance issues, in terms of speed and memory consumption, as all branched rays are naively recorded to perform backpropagation on them.
Application on volumetric real-world data with high resolution was thus almost prohibitive.

In this paper, we hence use an adapted version of Mitsuba 3 \cite{jakob2022mitsuba3} which is build on top of the DrJit autodiff framework \cite{Jakob2020DrJit}.
It is an advancement of Enoki and thus specifically tailored to efficiently handle the vast amount of branching, which naturally occurs in raytracing applications.

Finally, we strive to re-render the optimized data:
we will use -- as shown in the teaser -- a re-rendering pipeline to present the optimized data in new ways and other environments.
Based on given physical parametrizations, we want to be able to slice the specimen, re-light it, and even immerse it in other media.
Especially the latter allows for the utilization of Inherent Optical Properties (IOPs) of Ocean water \cite{preisendorfer1964physical,mobley2022oceanic}, which are frequently measured and publicly available (see, e.g, Jerlov Water Types \cite{solonenko2015inherent} as a prominent example).
An optimization run within the differentiable PBRT framework over time can be seen in Figs. \ref{fig:dp_opt}, \ref{fig:pf_opt} for our two captured datasets.

\newcommand{\ra}{\rightarrow}
\newcommand{\Ra}{\Rightarrow}

\newcommand{\atan}[1]{\mbox{atan2} \, ({#1})}

\newcommand  {\N}   {{\sf N}}

\renewcommand{\P}   {{\sf P}}

\newcommand  {\M}[1]{{\cal {#1}}}

\newcommand{\vd}{\:}

\newcommand{\vdot}{{\bf\cdot}}

\renewcommand{\d}[1]{\mbox{\boldmath$#1$}}
\newcommand{\di}[1]{\mbox{{\begin{footnotesize}\boldmath$#1$\end{footnotesize}}}}
\newcommand{\dii}[1]{\mbox{{\begin{scriptsize}\boldmath$#1$\end{scriptsize}}}}
\newcommand{\un}[1]{\underline{#1}}

\newcommand{\s}[1]{{\un{#1}}}

\newcommand{\sd}[1]{{\un{\mbox{\boldmath{$#1$}}}}}

\newcommand{\dd}[1]{{\mbox{\boldmath$#1$\unboldmath}}}

\newcommand{\est}[1]{\widehat{#1}}

\newcommand{\estd}[1]{\hat{\mbox{\boldmath$#1$}}}

\newcommand{\cs}[2]{\; ^{#1}\!\,{#2}}

\newcommand{\cv}[2]{\;^{#1}\!\,{\mbox{\boldmath$#2$}}}

\newcommand{\cq}[2]{\;^{#1}\!\,{\bf #2}}

\newcommand{\tra}[0]{{\sf T}}
\newcommand{\trans}[0]{^{\sf T}}
\newcommand{\transs}[0]{{'}^{\sf T}}
\newcommand{\transss}[0]{{''}^{\sf T}}
\newcommand{\transsss}[0]{{'''}^{\sf T}}
\newcommand{\transi}[0]{^{-\sf T}}

\newcommand{\transn}[0]{^{(0)\sf T}}
\newcommand{\transsn}[0]{{'}^{(0)\sf T}}
\newcommand{\transssn}[0]{{''}^{(0)\sf T}}

\newcommand{\sn}[0]{{'}^{(0)}}
\newcommand{\ssn}[0]{{''}^{(0)}}

\newcommand{\Amatrix}[0]{{\,\sf
		I\!\!\protect\rule[1.45ex]{0.8em}{0.8pt}\!\!\sf I \,}}

\newcommand{\Aqmatrix}[0]{\overline{\Amatrix}}
\newcommand{\Bqmatrix}[0]{\overline{\Bmatrix}}

\newcommand{\dual}[1]{{\overline{#1}}}

\newcommand{\tr}[0]{\sf T}

\newcommand{\q}[1]{{{\bf #1}}}

\newcommand{\supremum}[2]{{{\underset{#1}{\mathtt{sup}}\left[ #2 \right]}}}

\newcommand{\func}[2]{{{\mathtt {#1}\left[ #2 \right]}}}

\newcommand{\funca}[1]{{{\mathtt {#1}}}}

\newcommand{\img}[2]{{{\mathcal {#1}\left[ #2 \right]}}}

\newcommand{\imga}[1]{{{\mathcal {#1}}}}

\newcommand{\sq}[1]{{{\s{\bf #1}}}}

\newcommand{\qi}[1]{{\mbox{\tiny #1}}}

\newcommand{\m}[1]{{\mbox{{\fontencoding{T1}\sffamily{\itshape #1}}}}}

\newcommand{\mq}[1]{{\q #1}}

\newcommand{\qindex}[1]{{{\bf #1}}}

\renewcommand{\vec}[0]{\mbox{vec}}

\newcommand{\lb}[0]{\mbox{lb}}

\newcommand {\ftr}       {\circ\hspace{-3.5pt}-\hspace{-5pt}\bullet}

\newcommand{\trace}[0]{\mbox{tr}}

\newcommand{\rank}[0]{\mbox{rank} }

\newcommand{\Diag}[0]{\mbox{Diag}}

\newcommand{\matr}[2]{%
	\left( \begin{array}{*#1{c}}    #2   \end{array} \right) %
}

\newcommand{\dete}[2]{%
	\left| \begin{array}{*#1{c}}    #2   \end{array} \right| %
}

\newcommand{\vvector}[4]
{    \left(
	\begin{array}{c}
		{#1} \\ {#2} \\{#3} \\ {#4}
	\end{array}
	\right) }

\newcommand{\fvector}[5]
{    \left(
	\begin{array}{c}
		{#1} \\ {#2} \\{#3} \\ {#4}\\ {#5}
	\end{array}
	\right) }

\newcommand{\svector}[6]
{    \left(
	\begin{array}{c}
		{#1} \\ {#2} \\{#3} \\ {#4}\\ {#5}\\ {#6}
	\end{array}
	\right) }

\newcommand{\dvector}[3]
{    \left(
	\begin{array}{c}
		{#1} \\ {#2} \\{#3}
	\end{array}
	\right) }

\newcommand{\zvector}[2]
{    \left(
	\begin{array}{c}
		{#1} \\ {#2}
	\end{array}
	\right) }

\newcommand{\zdmatrix}[6]
{    \left(
	\begin{array}{ccc}
		{#1} & {#2} & {#3} \\
		{#4} & {#5} & {#6}
	\end{array}
	\right) }

\newcommand{\zvmatrix}[8]
{    \left(
	\begin{array}{cccc}
		{#1} & {#2} & {#3} & {#4}\\
		{#5} & {#6} & {#7} & {#8}
	\end{array}
	\right) }

\newcommand{\dzmatrix}[6]
{    \left(
	\begin{array}{ccc}
		{#1} & {#2} \\
		{#3} & {#4} \\
		{#5} & {#6}
	\end{array}
	\right) }

\newcommand{\dmatrix}[9]
{    \left(
	\begin{array}{ccc}
		{#1} & {#2} & {#3} \\
		{#4} & {#5} & {#6} \\
		{#7} & {#8} & {#9}
	\end{array}
	\right) }

\newcommand{\zmatrix}[4]
{    \left(
	\begin{array}{cc}
		{#1} & {#2} \\
		{#3} & {#4}
	\end{array}
	\right) }

\newcommand{\rot}[2]{\d{R}_{#1}(#2)}

\newcommand{\rott}[2]{\d{R}\trans_{#1}(#2)}

\newcommand{\zdet}[4]
{    \left|
	\begin{array}{cc}
		{#1} & {#2} \\
		{#3} & {#4}
	\end{array}
	\right| }

\newcommand{\ddet}[9]
{    \left|
	\begin{array}{ccc}
		{#1} & {#2} & {#3}\\
		{#4} & {#5} & {#6}\\
		{#7} & {#8} & {#9}
	\end{array}
	\right| }

\newcommand{\Gs}{{G_{\sigma}}}

\newcommand{\nab}[1]{\nabla \! #1}

\newcommand{\nabs}[1]{\nabla_{\!\!\sigma} #1}

\newcommand{\sg}[1]{\dd {\Gamma} \! #1}

\newcommand{\asg}[1]{\overline {\dd {\Gamma} \! #1}}

\newcommand{\assg}[1]{\overline {\dd {\Gamma}_{\!\!\sigma} #1}}

\newcommand{\curv}[1]{\kappa_\sigma #1}

\newcommand{\E}{{\rm e}}

\newcommand{\mB}{\rm I\!B}
\newcommand{\mC}{\mathchoice {\setbox0=\hbox{$\displaystyle\rm
			C$}\hbox{\hbox to0pt{\kern0.4\wd0\vrule height0.9\ht0\hss}\box0}}
	{\setbox0=\hbox{$\textstyle\rm C$}\hbox{\hbox
			to0pt{\kern0.4\wd0\vrule height0.9\ht0\hss}\box0}}
	{\setbox0=\hbox{$\scriptstyle\rm C$}\hbox{\hbox
			to0pt{\kern0.4\wd0\vrule height0.9\ht0\hss}\box0}}
	{\setbox0=\hbox{$\scriptscriptstyle\rm C$}\hbox{\hbox
			to0pt{\kern0.4\wd0\vrule height0.9\ht0\hss}\box0}}}
\newcommand{\mD}{\rm I\!D}
\newcommand{\mE}{\rm I\!E}
\newcommand{\mF}{\rm I\!F}
\newcommand{\mG}{\mathchoice {\setbox0=\hbox{$\displaystyle\rm
			G$}\hbox{\hbox to0pt{\kern0.4\wd0\vrule height0.9\ht0\hss}\box0}}
	{\setbox0=\hbox{$\textstyle\rm G$}\hbox{\hbox
			to0pt{\kern0.4\wd0\vrule height0.9\ht0\hss}\box0}}
	{\setbox0=\hbox{$\scriptstyle\rm G$}\hbox{\hbox
			to0pt{\kern0.4\wd0\vrule height0.9\ht0\hss}\box0}}
	{\setbox0=\hbox{$\scriptscriptstyle\rm G$}\hbox{\hbox
			to0pt{\kern0.4\wd0\vrule height0.9\ht0\hss}\box0}}}
\newcommand{\mH}{\rm I\!H}
\newcommand{\mI}{\rm I\!I}
\newcommand{\mJ}{\mathchoice {\setbox0=\hbox{$\displaystyle\rm
			J$}\hbox{\hbox to0pt{\kern0.4\wd0\vrule height0.9\ht0\hss}\box0}}
	{\setbox0=\hbox{$\textstyle\rm J$}\hbox{\hbox
			to0pt{\kern0.4\wd0\vrule height0.9\ht0\hss}\box0}}
	{\setbox0=\hbox{$\scriptstyle\rm J$}\hbox{\hbox
			to0pt{\kern0.4\wd0\vrule height0.9\ht0\hss}\box0}}
	{\setbox0=\hbox{$\scriptscriptstyle\rm J$}\hbox{\hbox
			to0pt{\kern0.4\wd0\vrule height0.9\ht0\hss}\box0}}}
\newcommand{\mk}{\rm I\!k}
\newcommand{\mK}{\rm I\!K}
\newcommand{\mL}{\rm I\!L}
\newcommand{\mM}{\rm I\!M}
\newcommand{\mN}{\rm I\!N}
\newcommand{\mO}{\mathchoice {\setbox0=\hbox{$\displaystyle\rm
			O$}\hbox{\hbox to0pt{\kern0.4\wd0\vrule height0.9\ht0\hss}\box0}}
	{\setbox0=\hbox{$\textstyle\rm O$}\hbox{\hbox
			to0pt{\kern0.4\wd0\vrule height0.9\ht0\hss}\box0}}
	{\setbox0=\hbox{$\scriptstyle\rm O$}\hbox{\hbox
			to0pt{\kern0.4\wd0\vrule height0.9\ht0\hss}\box0}}
	{\setbox0=\hbox{$\scriptscriptstyle\rm O$}\hbox{\hbox
			to0pt{\kern0.4\wd0\vrule height0.9\ht0\hss}\box0}}}
\newcommand{\mP}{\rm I\!P}
\newcommand{\mQ}{\mathchoice {\setbox0=\hbox{$\displaystyle\rm
			Q$}\hbox{\raise 0.15\ht0\hbox to0pt{\kern0.4\wd0\vrule
				height0.8\ht0\hss}\box0}}{\setbox0=\hbox{$\textstyle\rm Q$}\hbox{\raise
			0.15\ht0\hbox to0pt{\kern0.4\wd0\vrule height0.8\ht0\hss}\box0}}
	{\setbox0=\hbox{$\scriptstyle\rm Q$}\hbox{\raise 0.15\ht0\hbox
			to0pt{\kern0.4\wd0\vrule height0.7\ht0\hss}\box0}}{\setbox0=
		\hbox{$\scriptscriptstyle\rm Q$}\hbox{\raise 0.15\ht0\hbox
			to0pt{\kern0.4\wd0\vrule height0.7\ht0\hss}\box0}}}
\newcommand{\mR}{\rm I\!R}
\newcommand{\mS}{\mathchoice
	{\setbox0=\hbox{$\displaystyle     \rm S$}\hbox{\raise0.5\ht0\hbox
			to0pt{\kern0.35\wd0\vrule height0.45\ht0\hss}\hbox
			to0pt{\kern0.55\wd0\vrule height0.5\ht0\hss}\box0}}
	{\setbox0=\hbox{$\textstyle        \rm S$}\hbox{\raise0.5\ht0\hbox
			to0pt{\kern0.35\wd0\vrule height0.45\ht0\hss}\hbox
			to0pt{\kern0.55\wd0\vrule height0.5\ht0\hss}\box0}}
	{\setbox0=\hbox{$\scriptstyle      \rm S$}\hbox{\raise0.5\ht0\hbox
			to0pt{\kern0.35\wd0\vrule height0.45\ht0\hss}\raise0.05\ht0\hbox
			to0pt{\kern0.5\wd0\vrule height0.45\ht0\hss}\box0}}
	{\setbox0=\hbox{$\scriptscriptstyle\rm S$}\hbox{\raise0.5\ht0\hbox
			to0pt{\kern0.4\wd0\vrule height0.45\ht0\hss}\raise0.05\ht0\hbox
			to0pt{\kern0.55\wd0\vrule height0.45\ht0\hss}\box0}}}
\newcommand{\mT}{\mathchoice {\setbox0=\hbox{$\displaystyle\rm
			T$}\hbox{\hbox to0pt{\kern0.3\wd0\vrule height0.9\ht0\hss}\box0}}
	{\setbox0=\hbox{$\textstyle\rm T$}\hbox{\hbox
			to0pt{\kern0.3\wd0\vrule height0.9\ht0\hss}\box0}}
	{\setbox0=\hbox{$\scriptstyle\rm T$}\hbox{\hbox
			to0pt{\kern0.3\wd0\vrule height0.9\ht0\hss}\box0}}
	{\setbox0=\hbox{$\scriptscriptstyle\rm T$}\hbox{\hbox
			to0pt{\kern0.3\wd0\vrule height0.9\ht0\hss}\box0}}}
\newcommand{\mU}{\mathchoice {\setbox0=\hbox{$\displaystyle\rm
			U$}\hbox{\hbox to0pt{\kern0.4\wd0\vrule height0.9\ht0\hss}\box0}}
	{\setbox0=\hbox{$\textstyle\rm U$}\hbox{\hbox
			to0pt{\kern0.4\wd0\vrule height0.9\ht0\hss}\box0}}
	{\setbox0=\hbox{$\scriptstyle\rm U$}\hbox{\hbox
			to0pt{\kern0.4\wd0\vrule height0.9\ht0\hss}\box0}}
	{\setbox0=\hbox{$\scriptscriptstyle\rm U$}\hbox{\hbox
			to0pt{\kern0.4\wd0\vrule height0.9\ht0\hss}\box0}}}
\newcommand{\mZ}{{\sf Z\hspace{-1.8ex}Z}}

\newcommand{\mone}{\mathchoice {\rm 1\mskip-4mu l} {\rm 1\mskip-4mu l}
	{\rm 1\mskip-4.5mu l} {\rm 1\mskip-5mu l}}

\section{Approach}
\subsection{Image Acquisition}\label{subsec:image_acquisition}
\begin{figure}[t]
	\begin{center}
		\includegraphics[width=.99\linewidth]{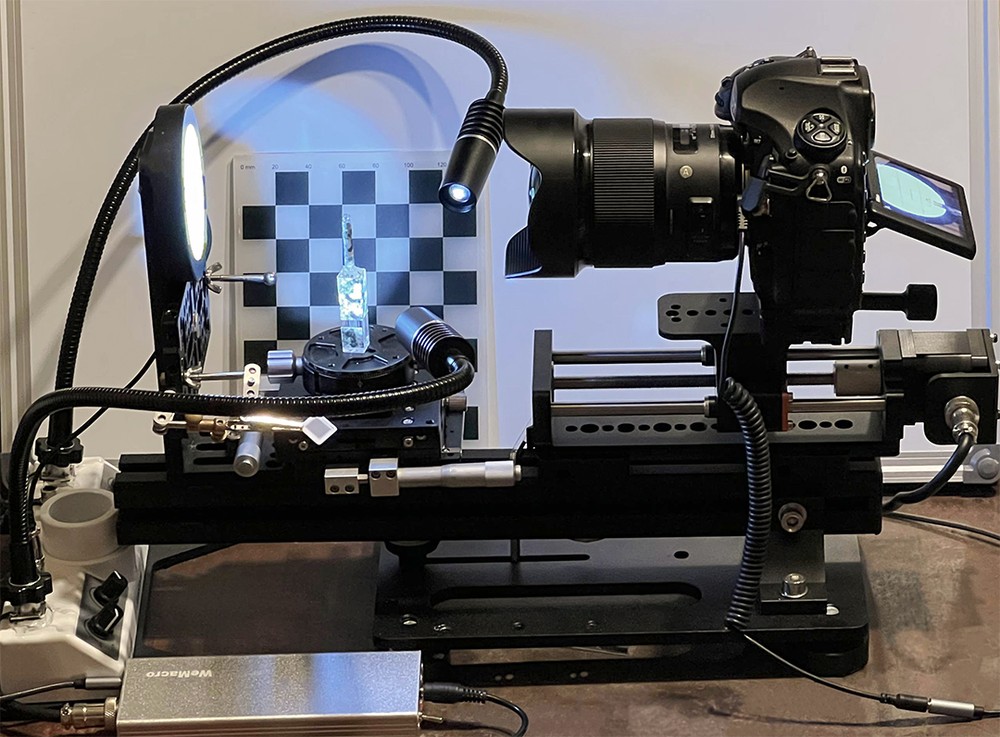}
	\end{center}
	\caption{Image acquisition setup: a macro camera takes brightfield images of small, partially translucent specimens.}
	\label{fig:image_capture_setup}
\end{figure}
\begin{figure}[t]
	\begin{center}
		\includegraphics[width=.47\linewidth]{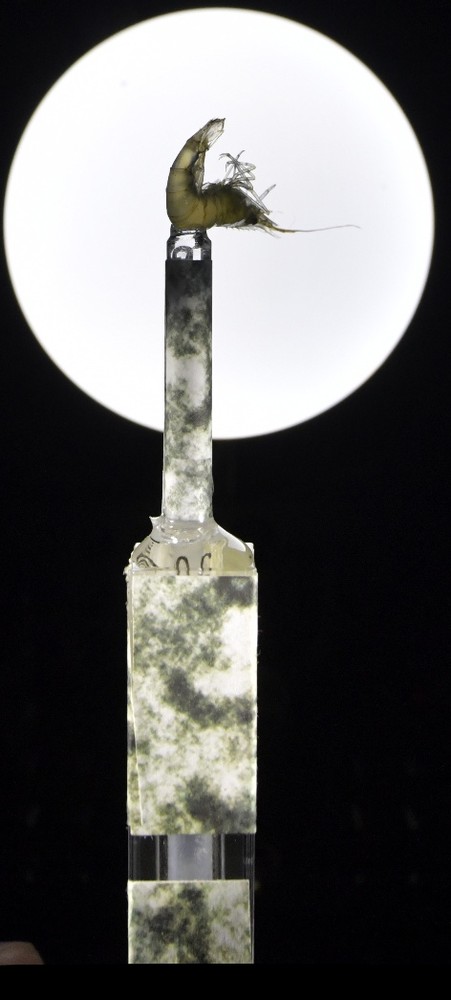}
		\includegraphics[width=.47\linewidth]{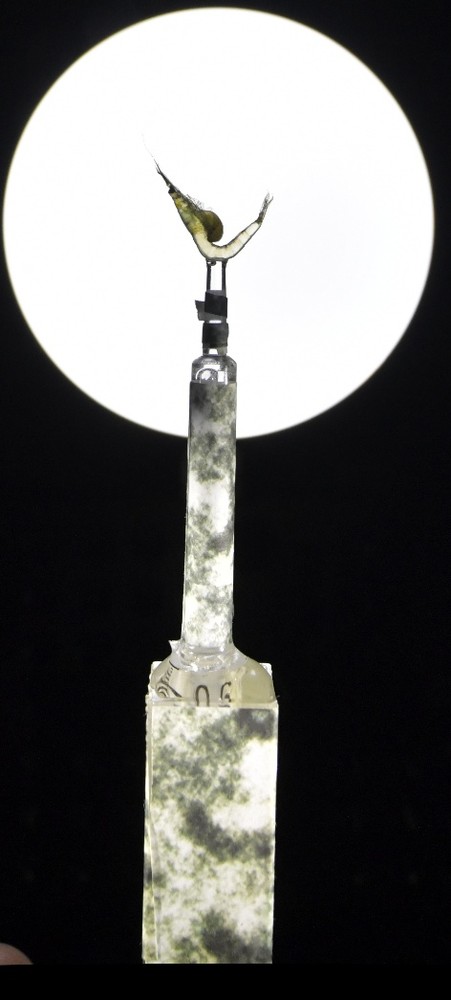}
	\end{center}
	\caption{Full image acquisition setup: The specimen is captured in front of an uncalibrated off-the-shelf microscope lightsource, on top of a base, equipped with random dot markers \cite{li2013multiple} (Left: Dwarf Prawn, Right:  Praunus Flexuosus).}
	\label{fig:full_img_capture}
\end{figure}

The full image acquisition setup is shown in Fig. \ref{fig:image_capture_setup}.
A SIGMA 20mm ART lens is mounted on a Nikon D850 DSLR camera.
A micro-rail driven by a stepping motor is used for focusing.
A base with random dot markers \cite{li2013multiple} is placed on a rotating platform, to foster the image registration and camera pose estimation in the sparse reconstruction step.
For capturing, the specimen is put on top of the base.
A rotating platform mounted on a xy-micro-stage is used to acquire images of objects from different viewing-directions.
The xy-micro-stage ensures that the object remains in the camera's field of view.
Two 6500K microscope gooseneck light sources are used to illuminate the markers below the object.
An uncalibrated off-the-shelf circular microscope light source is used for brightfield illumination.
During image acquisition, the camera is line triggered and uses a "quiet" mode to reduce image blur caused by vibration of the mechanical shutter.
The acquisition parameters are: ISO$100$, exposure time $300$ms, and the full image size is $8256\times5504$px.
Two non-cropped example images from our datasets are depicted in Fig \ref{fig:full_img_capture}.

\subsection{Pose Estimation / SfM}
To determine the poses of the cameras, we use the Structure-from-Motion (SfM) as implemented by COLMAP \cite{schoenberger2016sfm,schoenberger2016mvs}.
The semi-transparent nature of the plankton introduces challenges for the traditional SfM process, that assumes opaque surfaces with Lambertian reflectance and well-structured textures.
To address this, we create robust markers that exhibit distinct features at various scales, which we affix to the base of the plankton object to be reconstructed (see lower part of Fig. \ref{fig:full_img_capture}). This approach enables the identification and matching of reliable features, thereby enhancing the robustness and accuracy of the subsequent 3D reconstruction process.
Our inspiration for this marker-based approach is drawn from the work of \cite{li2013multiple}, where similar patterns were employed for multi-camera calibration.
Afterwards, the camera poses will be used for a differentiable raytracer (and a NeRF), as such gradient-based approaches only allow for pose-refinement and not the determination of the poses without any prior information.

\subsection{Optional Initialization from Emissive Volume}\label{sec:nerf_init}
\begin{figure}[t]
	\begin{center}

		\includegraphics[trim=50 0 50 50,clip,width=.325\linewidth]{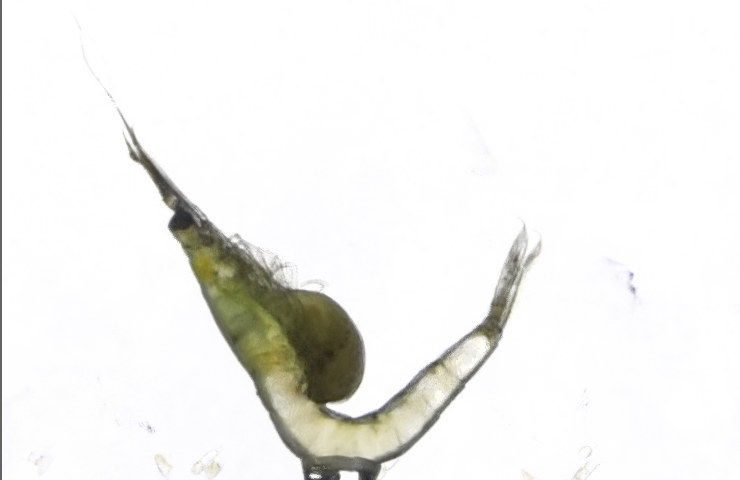}
		\includegraphics[trim=50 0 50 50,clip,width=.325\linewidth]{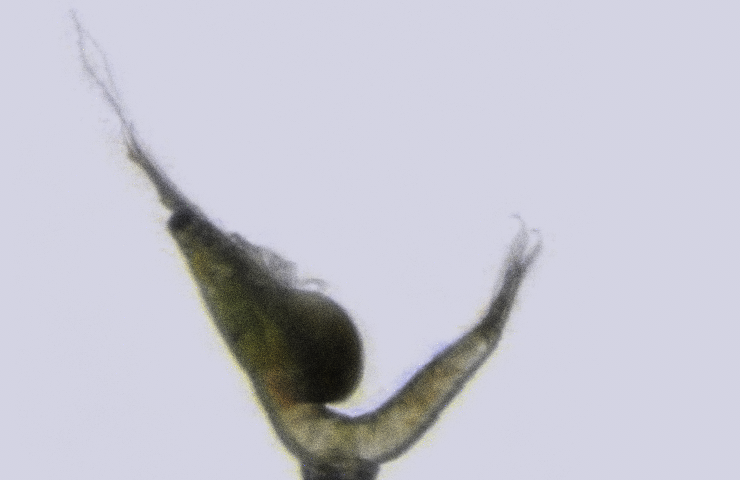}
		\includegraphics[trim=50 0 50 50,clip,width=.325\linewidth]{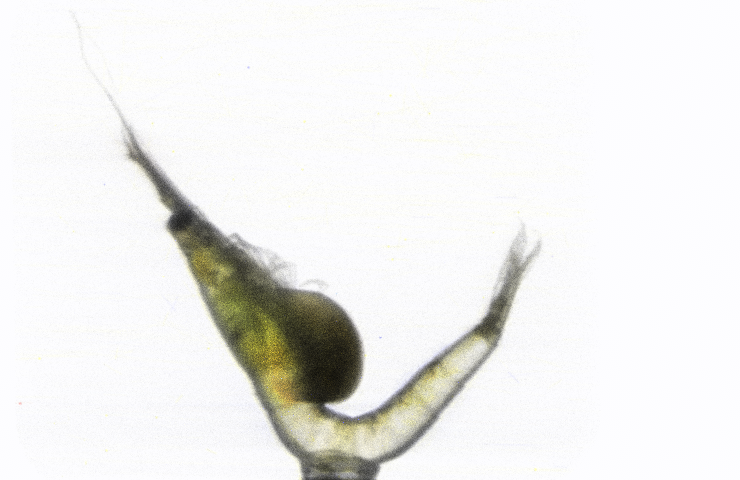}
	\end{center}
	\caption{From left to right:  optimization result in NeRF, density extracted, as described in Sect. \ref{sec:nerf_init} and re-rendered in \textbf{MO} condition with 0 iterations (result shown with 128 spp), refined for 15 iterations in \textbf{MO+LO+NH} condition (result shown with 128spp).}
	\label{fig:nerf_init}
\end{figure}
We know that the density optimization of an emissive volume often behaves more stable than the physically based one.
And on top of that, density initialization can be used to dodge local minima in physically-based medium optimizations \cite{nimierdavid2022unbiased}.
Hence, we first train a Nerfacto-Huge model \cite{nerfstudio} of the plankton object as an initialization step.
As in the original NeRF model \cite{mildenhall2021nerf}, we compute the expected color $\hat{C}(\d r)$\footnote{Throughout the paper, we will typeset  scalars as $s$, Euclidean vectors as $\d v$, and matrices as $\m m$.} of any camera ray $\d r(t) = \d o + t \d d$, with camera center $\d o$ and viewing direction $\d d$ from the volume:
\begin{equation}\label{eq:nerf_int}
	\hat{C}(\d r) = \int_{t_n}^{t_f} T(t)\sigma(t)\d c(t, \d d) dt
\end{equation}
where integration along the ray is bounded in the range of $t \in [t_n, t_f]$; $\sigma(t)$ and $\d c(t, \d d)$ represents the density and its emitted color towards the camera center at the point $\d r(t)$; $T(t)$ represents the accumulated transmittance along the ray from $t_n$ to $t$, and is given by:
\begin{equation}\label{eq:nerf_trans}
	T(t) = \exp \left( -\int_{t_n}^{t} \sigma(s)ds \right).
\end{equation}
To evaulate Eqs. \ref{eq:nerf_int}, \ref{eq:nerf_trans} we need to train a mapping from position and viewing direction to emitted color and density in form of a MLP:
\begin{equation}\label{eq:nerf_mlp}
	F_{\d \Theta} : (t,\d d) \rightarrow (\d c,\sigma).
\end{equation}
Finally, the parameters $\d \Theta$ of the Nerfacto-Huge model are optimized by minimizing the image reconstruction loss over single rays.
Since each point within the volume emits a certain amount of light, it is essentially a non-physical emissive volume.
We now strive to extract its density and map it to the physically-based medium to subsequently refine it in the differentiable raytracer as depicted in Fig. \ref{fig:nerf_init}.
To map the one-dimensional density $\sigma$ to the three-dimensional density ${\d \sigma_t}$ over all wavelenghts, we pursue the following line of thought:
We assume that in our scenario, the emissive medium locally behaves inverted to a physically-based one: the first emits light over the spectrum, while the second attenuates over the whole spectrum. Then, we can further assume in the wavelength-dimension that the attenuated light is the inverse of the emitted light.

Based on this reasoning, we can define the following extraction rule: we simply take the mean local light emittance over $n$ directional samples and substract that number from the light's radiance  $\mathbb{\d L} \in \mathbb{R}^3$ as estimated by the differentiable raytracer.
As this is an optional step, a stable light estimate can be extracted from a run without medium intialization. The \textit{inverse emittance} can be used to distribute the one-dimensional density into the three channels like so:
\begin{equation}\label{eq:inv_em}
	\d \sigma_t = \sigma(t)  \left(\mathbb{\d L} - \frac{1}{n} \sum\limits_{\theta} \sum\limits_{\phi} \d c(t, \d d)\right),
\end{equation}
where $\theta,\phi$ are the spherical angles encoding the directions $\d d$ we sample from.

In addition, we use a triangle mesh, extracted from the NeRF, to serve as a 3D medium boundary between the optimized heterogeneous medium and the parameterized homogeneous medium, parametrized by real-world-measurements.
Hence, we can re-render the volumetric models in other volumes, whose parameters are given by the measurement environment (see teaser image).

\subsection{Differentiable Physically-Based Raytracing}\label{sec:DPBRT}
To enable the optimization of implicit representations of the specimens in the framework of Differentiable Physically-Based Raytracing (DPBRT), we rely on a heterogeneous medium.
It is modelled as a voxel grid of local isotropic medium parameters, which are  queried in an interpolated 3D lookup during the rendering process.
We denote the three dimensions of the voxel grid as $X$, $Y$ and $Z$.

In our approach, the homogeneous medium is described by its attenuation and scattering properties over the three considered wavelengths.
The former parameter describes the combined loss of radiance due to absorption $\d \sigma_a$ and out-scattering $\d \sigma_s$.
In an isotropic medium the \textit{attenuation} is independent of the incoming direction and and interaction point, hence the voxelized heterogeneous version is
\begin{equation}
	\d \sigma_a + \d \sigma_s = \d \sigma_t \in \mathbb{R}^{3\times X\times Y \times Z}.
\end{equation}
The \textit{single scattering albedo} gives the local probability of out scattering, given by the ratio of the scattering coefficient and the attenuation:
\begin{equation}
	\frac{\d \sigma_s}{\d \sigma_t}= \d \rho_a \in \mathbb{R}^{3\times X\times Y \times Z}.
\end{equation}
Finally, if a scattering event occurs, the scattering phase function describes a distribution of directions, which can be sampled from.
We use the classical Heney Greenstein function \cite{henyey1941diffuse}, which was designed to provide a good fit to measured data \cite{pharr2023physically}
\begin{equation}
	p_{HG}(cos \theta) = \frac{1}{4 \pi} \frac{1 -g^2}{(1+g^2+2g(\cos \theta))^{3/2}}.
	\label{eq:phg}
\end{equation}
Due to the isotropy assumption, also the phase function is locally independent of direction and position of the scattering event.
Hence, we can describe the scattering event exhaustively by the angle between the incoming- and the outgoing light ray.
In a heterogeneous grid, we optimize a vector of local phase parameters
\begin{equation}
	\d g \in \mathbb{R}^{X\times Y \times Z}.
\end{equation}
Note that here, parameter $g$ does not vary with wavelengths, which means that we can get a different amount of scattering per wavelength but that the shape of the phase function will remain the same over the whole spectrum.
Finally, to give our model the necessary flexibility to represent the entire scene: we parametrize the light's radiance as $\mathbb{\d L} \in \mathbb{R}^3$,
and optimize it as well.
Thus, we account for the combined effect of the camera's properties (e.g., white-balance, spectral-response) and the actual radiance itself.

The model described above provides a sufficient informational basis to solve the Radiative Transfer Equation (RTE) \cite{subrahmanyan1960radiative,arvo1993transfer} to yield a radiance at point $\d r$ in direction $\d d$:
\begin{equation}{\small
		L(\d r,\d d) = T(\d r_0, \d r) L_0(\d p_0,-\d d)+\int_{0}^{t}  T(\d r', \d r) L_s(\d r', \d -d)   d t'}
\end{equation}
where $\d p_0$ is a point on a surface whose emitted radiance $ L_0(\cdot)$ can be computed based on $\mathbb{\d L}$,
$L_s(\cdot)$ is the source term, which accounts for radiance accumulated along the ray,
furthermore, the points along the ray are given by $\d r' = \d r + t' \d d$,
and finally, $T(\cdot)$ denotes the transmission and can be efficiently estimated in the heterogeneous case by ratio tracking \cite{novak2014residual}.
The source term $L_i(\cdot)$ can be evaluated by recursively importance sampling new ray directions wrt. the phase function, and subsequently sampling the free flight distance with differentiable delta tracking \cite{Vicini2021PathReplay}.
For optimization, we use a volumetric Monte-Carlo integrator.
Specifically, we rely on Mitsubas \textit{prbvolpath}-integrator, which implements Path Replay Backpropagation \cite{Vicini2021PathReplay}, which in turn builds on top of Radiative Backpropagation \cite{NimierDavid2020Radiative}.
The combination of the latter ideas allows for a linear time behaviour and constant memory consumption while backpropagating  gradients through DPBRT-algorithms.

\subsection{Re-Rendering}
\begin{figure*}[t]
	\begin{center}
		\includegraphics[width=1.\linewidth]{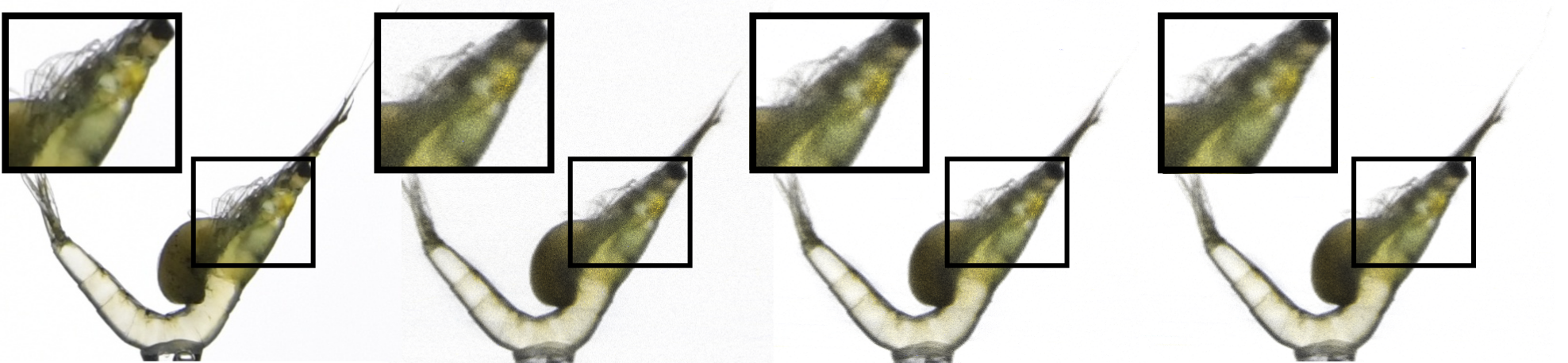}
	\end{center}
	\caption{Improvements in re-rendering pipeline, from left to right: Errors wrt. GT-image, final optimization image at 128 spp (\textbf{MSE}: 71.57, \textbf{PSNR}: 29.58, \textbf{SSIM}:  0.8620), re-rendered image with \textit{pbrvolpath} at 2048 spp 128 spp (\textbf{MSE}: 58.67 , \textbf{PSNR}: 30.44, \textbf{SSIM}: 0.9458 ), re-rendered image with \textit{volpathmis} at 2048 spp 128 spp (\textbf{MSE}: 57.42 , \textbf{PSNR}: 30.53, \textbf{SSIM}:  0.9480). Please note, the decreased error-level stems from an evaluation on a single training set image.}
	\label{fig:re-renderingpipeline}
\end{figure*}
For re-rendering the data, we simply use more samples per pixel (SPP), to yield better images from the same data-base.
In addition, the application of Mitsubas \textit{Volpathmis} integrator, which uses null-scattering in a path integral formulation to minimize sample variance by employing multiple importance sampling in the wavelength domain of participating media \cite{miller19null}, is also advantageous (see Fig. \ref{fig:re-renderingpipeline}).

\begin{figure}[ht]
	\begin{center}
		\includegraphics[trim={0 0 15cm 0}, clip, width=.48\linewidth]{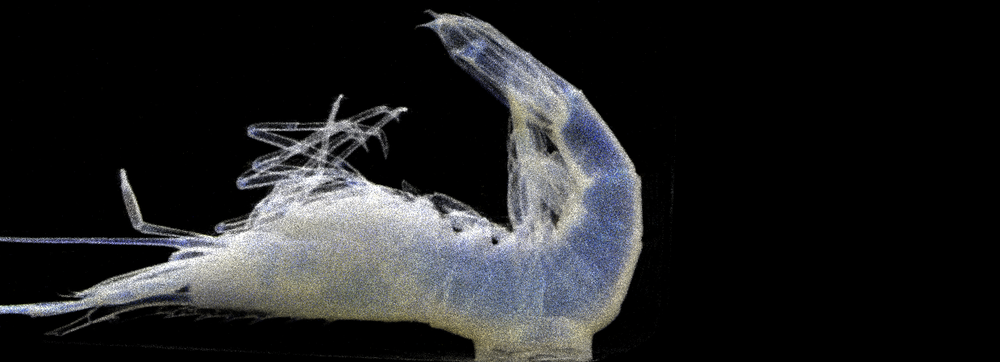}
		~~\includegraphics[trim={0 0 15cm 0}, clip, width=.48\linewidth]{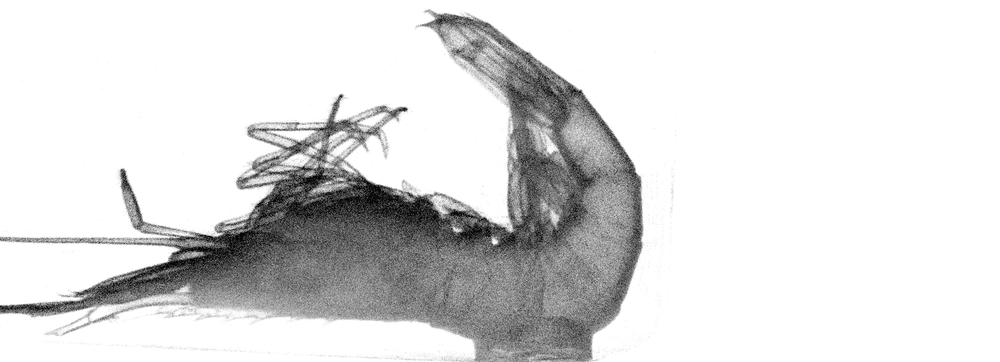}
	\end{center}
	\caption{Re-lighting of the optimized models; Left: darkfield mode -- i.e., with a light source above the medium and Right: inverse darkfield as used by the UVP-series plankton sensors. }
	\label{fig:re_light}
\end{figure}

\begin{figure}[ht]
	\begin{center}
		\includegraphics[width=1\linewidth]{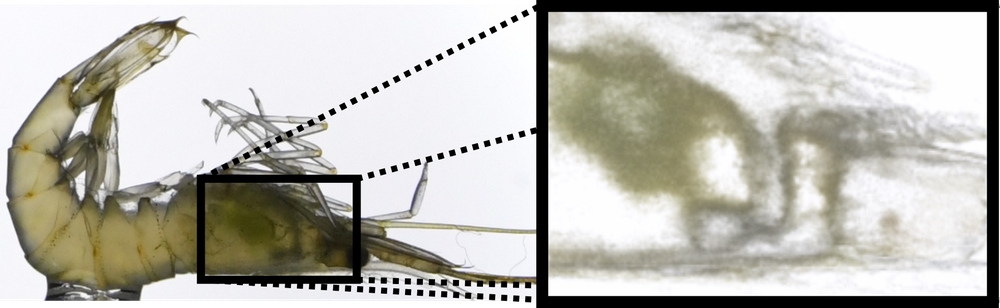}
	\end{center}
	\caption{Sliced re-rendering of the optimized Dwarf Prawn model, to get a glimpse of the interior structures.}
	\label{fig:slice}
\end{figure}

As our optimized model is completely disentangled from any environmental properties, we can use the full variety of image synthetization options available in the PBRT-realm in arbitrary environmental conditions.
We use this flexibility to \textit{(i)} synthesize sharper images with less noise as compared to the optimization process  (see Fig. \ref{fig:re-renderingpipeline}), \textit{(ii)} model different illumination scenarios as induced by different measurement approaches (see Fig. \ref{fig:re_light}), \textit{(iii)} slice, rotate and re-render to gain a better view inside the specimen (see Fig. \ref{fig:slice}),
and finally \textit{(iv)} introduce new measurement environments (e.g., underwater) based on actual measured physical properties (see teaser image).


\section{Evaluation}\label{sec:evaluation}

\subsection{Datasets}\label{sec:datasets}

\begin{figure*}[t]
	\begin{center}
		\includegraphics[width=.095\linewidth]{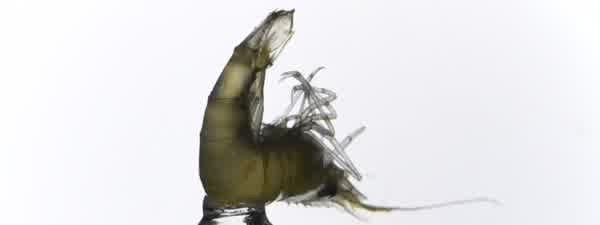}
		\includegraphics[width=.095\linewidth]{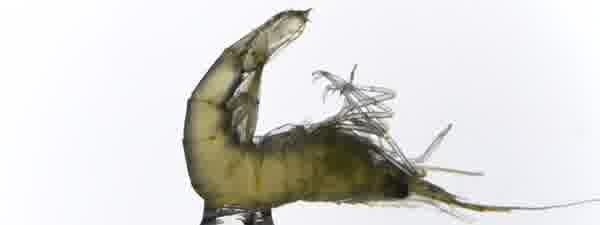}
		\includegraphics[width=.095\linewidth]{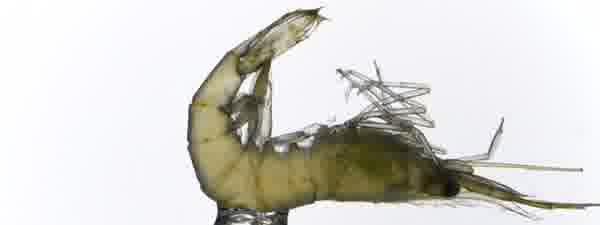}
		\includegraphics[width=.095\linewidth]{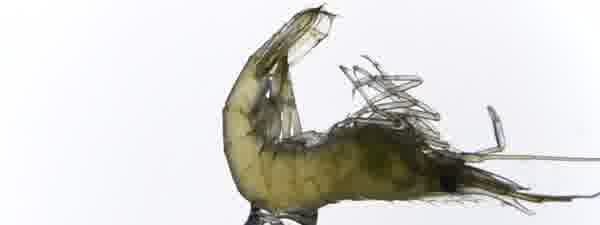}
		\includegraphics[width=.095\linewidth]{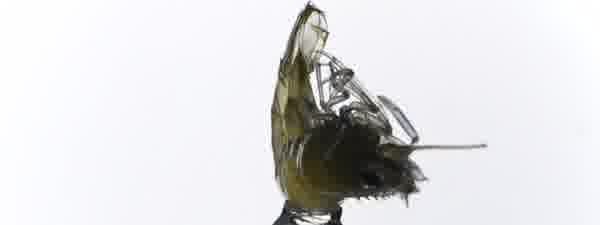}
		\includegraphics[width=.095\linewidth]{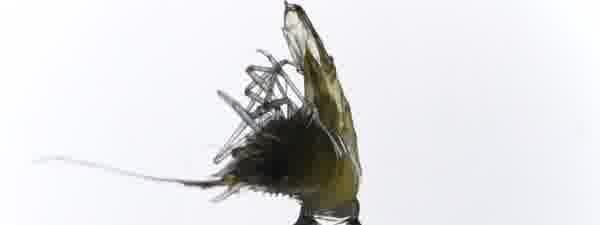}
		\includegraphics[width=.095\linewidth]{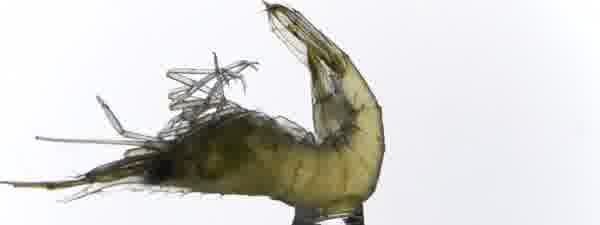}
		\includegraphics[width=.096\linewidth]{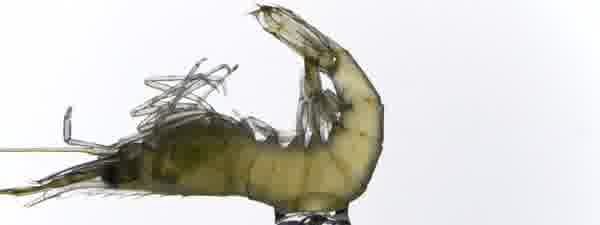}
		\includegraphics[width=.095\linewidth]{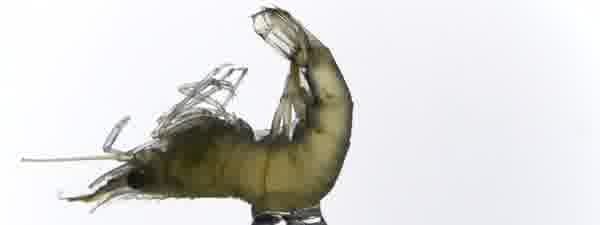}
		\includegraphics[width=.095\linewidth]{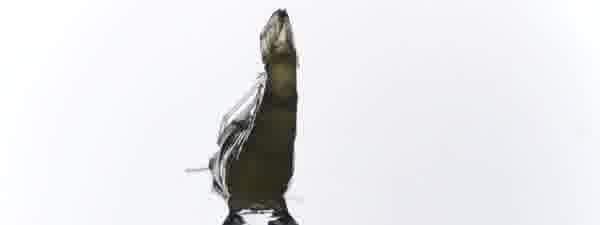}\\~\\
		\includegraphics[width=.095\linewidth]{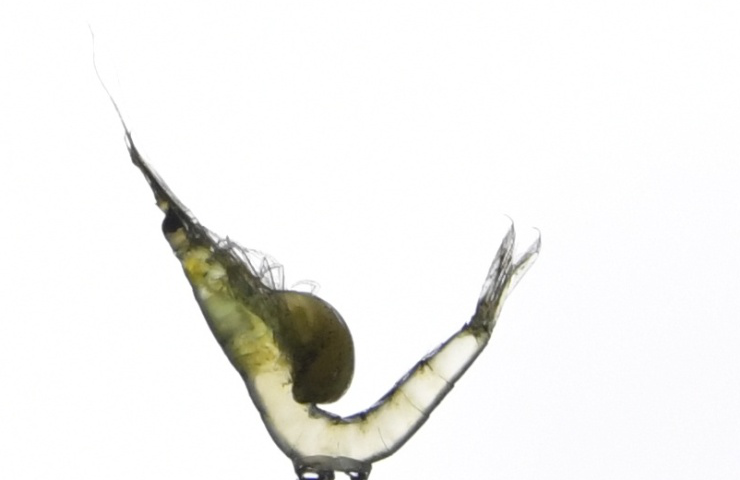}
		\includegraphics[width=.095\linewidth]{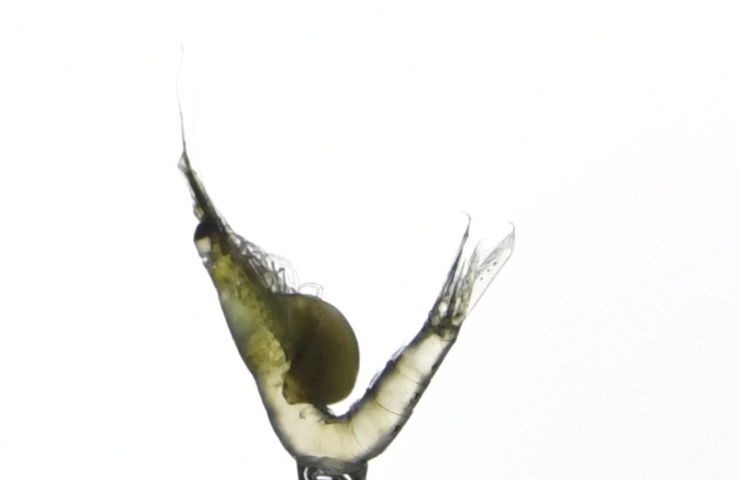}
		\includegraphics[width=.095\linewidth]{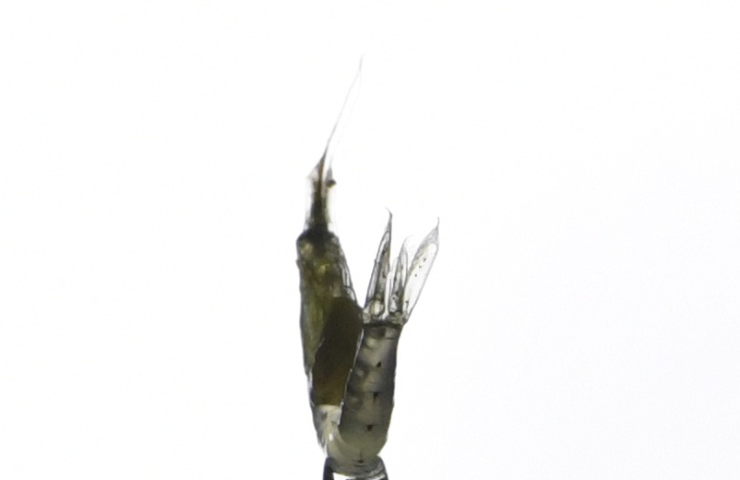}
		\includegraphics[width=.095\linewidth]{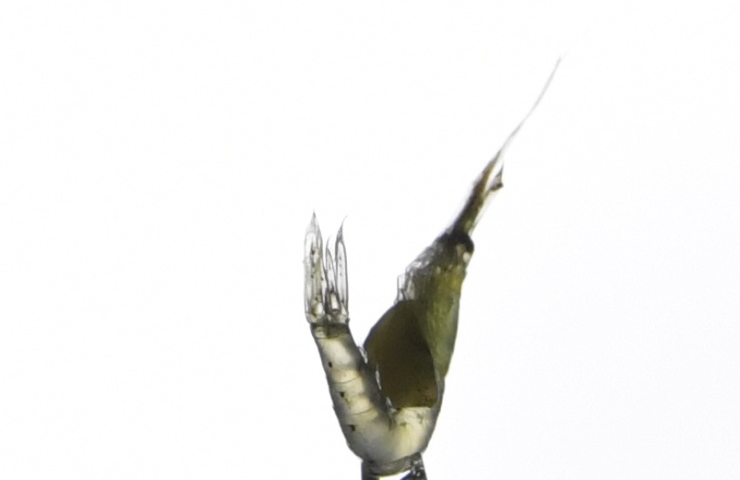}
		\includegraphics[width=.095\linewidth]{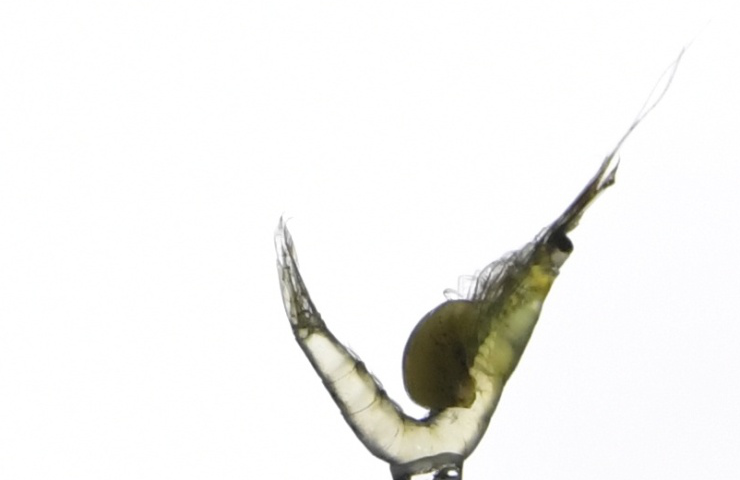}
		\includegraphics[width=.095\linewidth]{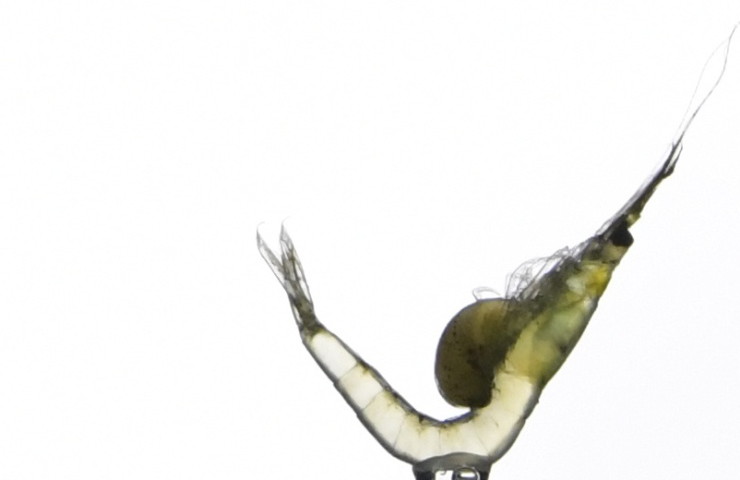}
		\includegraphics[width=.095\linewidth]{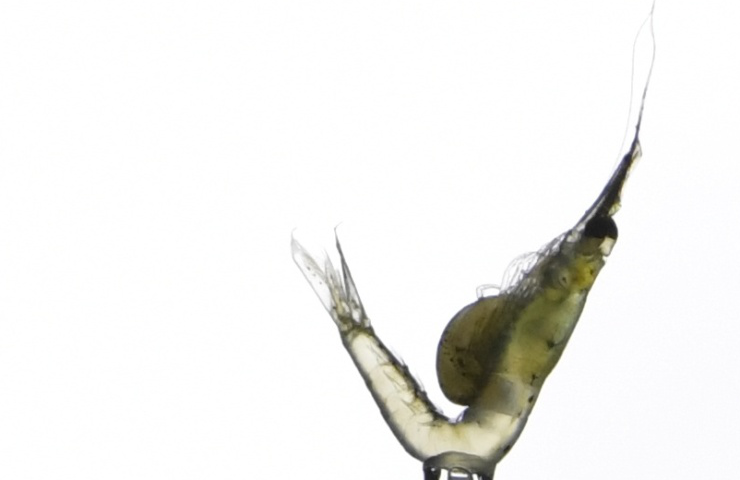}
		\includegraphics[width=.095\linewidth]{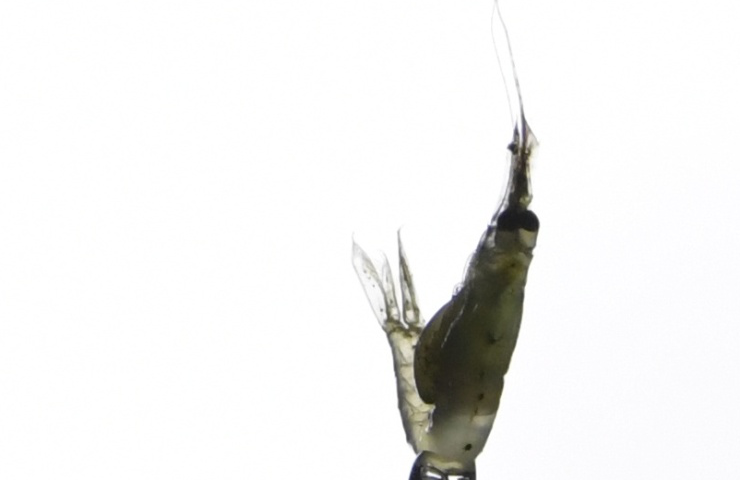}
		\includegraphics[width=.095\linewidth]{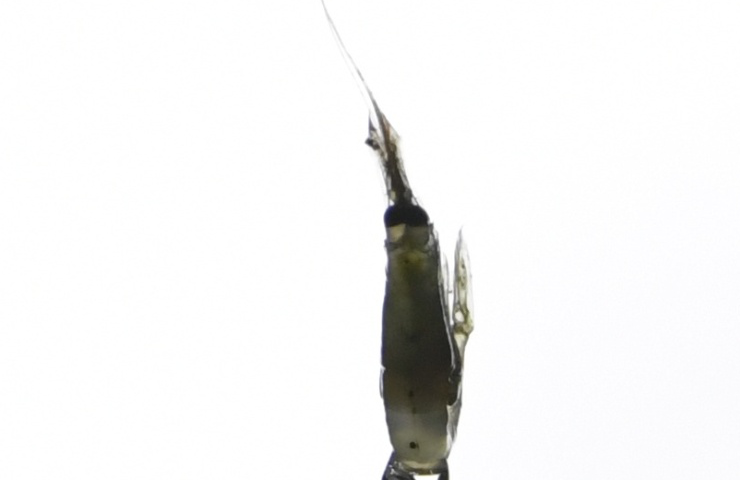}
		\includegraphics[width=.095\linewidth]{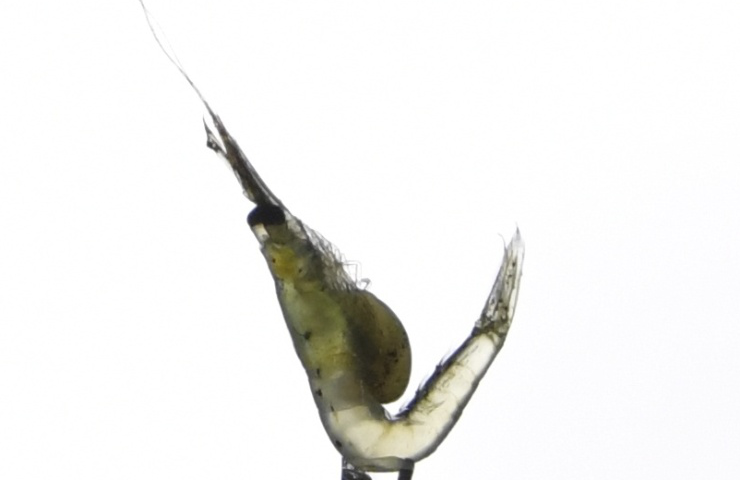}
	\end{center}
	\caption{Cropped ($1240\times450$px) training subset for Dwarf Prawns (1st row) and Praunus Flexuosus (2nd row), which both contain 48 images in total. Full images and their cropping will be shown in supplementary material.}
	\label{fig:dataset_dp}
\end{figure*}

We recorded two brightfield datasets with the method described in Sect. \ref{subsec:image_acquisition}.
The first row in Fig. \ref{fig:dataset_dp} depicts the Dwarf Prawn \textbf{DP} with a body length of $\sim 27.8 mm$, while the second presents scans of a Praunus Flexuosus \textbf{PF} with a body length of $\sim 17.4 mm$.
As we have limited time to capture the specimen, we took a training set with 48 images and a validation set of the same size but with pose-offsets in all six degrees of freedom.
For evaluation, the datasets were cropped, such that only the illuminated background can be seen (c.f.  Fig. \ref{fig:dataset_dp} to Fig. \ref{fig:full_img_capture} ).
This ensures a fair comparison between the NeRF and our approaches, as an explicitly visible moving light source breaks the assumptions underlying NeRFs and consequently yields a lot of floaters.

\subsection{Setup} \label{sec:eval:setup}
We run different conditions in a combined ablation and comparison to the Nerfacto-Huge model (\textbf{NH}) study.
In the  \textbf{MO} condition, \textit{all medium parameters} are optimized while the the lights radiance is fixed.
With added \textit{light optimization}, the condition is denoted as \textbf{MO+LO}.
And finally, in the \textbf{MO+LO+NH} condition, the medium is \textit{initialized with the density} from \textbf{NH} and \textit{light optimization} is conducted.

We use a simple multiview mean squared error loss function on the images, to determine the photometric error
between the reference and the synthesized images.

In all conditions involving \textbf{MO}, we employ a sparse variant of the Adam optimizer \cite{kingma2014adam}.
In this variant, only moments that show up in the gradient get updated, and only those portions of the gradient get applied to the parameters.
In all conditions, we use an adaptive learning rate over 60 iterations, if not stated otherwise.
Three steps are applied in the settings \textbf{MO}, and \textbf{MO+LO}, where we start the warmup-phase with 1e-3, in iteration 10 switch to 2e-4, and finally refine with 55e-6 from iteration 20 on.
For the \textbf{MO+LO+NH} on the \textbf{PF}-data, we iterated 15 epochs and used an initial learning rate of 75e-5 and then went down to 1e-5 after 5 iterations.
For the \textbf{MO+LO+NH} on the \textbf{DP}-data, we iterated 15 epochs and used an initial learning rate of 1e-3 and then went down to 1e-4 after 3 iterations.

Throughout all conditions, we use 128 samples per pixel (spp) to synthesize the single images.
All volume grids are initialized uniformly like so: $ \d \sigma_t$ is initialized with 0, $\d \rho_a$ with .5 and $\d g$ with 0 which induces an isotropic scattering. The resolution of the volume grid on the \textbf{DP}-data is $x=390, y=315, Z=960$, while the resolution  of the volume grid on the \textbf{PF}-data is set as $x=675, y=225, Z=729$.

In all conditions, the light's radiance is initialized according to the manufacturers color temperature label.
We observed, that it is necessary to optimize the light's radiance to account for the combined effect of the camera (whitebalance, response curve, etc) and the light's radiance. However, optimizing the light throughout all steps yields and oscillating behaviour between the light and the medium. Hence, in all settings where the radiance is optimized, we stop its optimization after 8 iterations.

We tuned the \textbf{Nerfacto-Huge}-model to yield the best possible result:
we set the \textit{learning rate} to 1e-2, with an exponential decay scheduler, the final learning rate is 1e-4. The \textit{max-num-iterations} is set to 100000.
In addition, we disable scene contraction, which is only meaningful when large scenes have to be embedded in the MLP and set the \textit{initial background color} to white.
As the unit of the density is reciprocal distance ($m^{-1}$), we scale the camera center to match the real physical size of the plankton object. Similar to scaling structure from motion results to real-world-units.
This is important, to get correctly scaled numbers when the density is extracted to be used in physically-based media.

\subsection{Results}
\begin{table}[ht]
	\centering
	\begin{tabular}{l|cccc}
		&   \textbf{MO} &  \textbf{MO+LO} & \textbf{NH}  & \textbf{MO+LO+NH}\\ \hline %
		$\downarrow$ MSE    & 1637.25 & 185.60 & 298.83 & \textbf{181.25} \\ %
		$\uparrow$ PSNR     & 15.99   & 25.57  & 23.76  & \textbf{25.68}\\  %
		$\uparrow$ SSIM    & 0.90    & 0.92   & 0.92   & \textbf{0.93} \\  %
	\end{tabular}
	\caption{Praunuous Flexuosus errors over 48 validation images.}
	\label{tab:pf_errors}
\end{table}

\begin{table}[ht]
	\centering
	\begin{tabular}{l|cccc}
		&   \textbf{MO}   &  \textbf{MO+LO} & \textbf{NH} & \textbf{MO+LO+NH} \\\hline
		$\downarrow$ MSE &  962.95 & 259.74 & 360.96  &   \textbf{244.42} \\
		$\uparrow$ PSNR &  18.30 & 24.01 & 22.84  & \textbf{24.35} \\
		$\uparrow$ SSIM &  0.87 & 0.87 & 0.87   & \textbf{0.89} \\
	\end{tabular}
	\caption{Dwarf Prawns: errors over 48 validation images.}
	\label{tab:dp_errors}
\end{table}

\begin{table}[ht]
	\centering
	\begin{tabular}{l|cccc}
		&  \textbf{MO}$^*$     &  \textbf{MO+LO}$^*$ & \textbf{NH}$^\dagger$ & \textbf{MO+LO+NH}$^{*\dagger}$ \\\hline
		\textbf{PF}      & 8.91 & 7.14 & \textbf{4.40}  & 4.40 + 6.18 \\
		\textbf{DP}     & 14.86 & 12.38& \textbf{4.41}  & 4.41 + 11.15 \\
	\end{tabular}
	\caption{Training run time in [h] on a ($^*$ on a RTX A6000 GPU,$~^\dagger$ on a GeForce RTX 3090 GPU).}
	\label{tab:runtimes}
\end{table}
\noindent
In Tab. \ref{tab:pf_errors} the mean errors of Praunuous Flexuosus validation set over 48 images are shown.
The differentiable renderer clearly needs the light optimization (compare \textbf{MO} against \textbf{MO+LO} over both sets), to optimize a noise free but also structurally correct image.
The \textbf{MO+LO} optimization clearly performs better than the NeRF (\textbf{NH}), which can be explained by a higher novel-view stability but a slightly less detailed re-rendering.
Finally, the  \textbf{MO+LO+NH} condition performs the best on all errors by combining high details with a big novel-view stability.
In Tab. \ref{tab:dp_errors}. the mean errors of the Dwarf Prawns  validation set over 48 images are shown.
Again, our approach \textbf{MO+LO} outperforms the NeRF \textbf{NH}, which already performs on a high level.
Finally, the initialization from NeRF can also further improve result over all measures, see  \textbf{MO+LO+NH}.

The computation times are stated in Tab \ref{tab:runtimes}.
The NeRF certainly exhibits the best runtime over both datasets.
Our method is highly impacted by the higher density of the Dwarf Prawn sample.
Furthermore, running less optimization variables does not lead to shorter computation times, but can even have the inverse effect (compare \textbf{MO} to \textbf{MO+LO} on both datasets).
This effect might be explained by a better distributed density, which can be traversed faster by the rays.
Finally, the combination of our approach with the NeRF does not seem to be faster than the \textbf{MO+LO} condition per se.\FloatBarrier

\section{Conclusion}
We have presented the first approach to reconstruct volumetric models of translucent, plankton-like objects. The resulting representation is suitable for being relighted and re-immersed into a different medium such as water. On top of being more flexible for our future intended use of relighting and submerging, the overall pipeline achieves better accuracy as compared to the state-of-the-art NeRF system Nerfacto when rendering previously unseen poses. Optimization of the light source parameters is key to obtaining good results with real-world data. The best results are obtained when using NeRF as initialization, presumably because of its stable density estimate. This scalar value however, needs to be disentangled to spectral/RGB parameters for PRBT rendering, which we proposed to solve by integrating NeRF responses at the local voxel over all directions. As this is work in progress, we intend to scan more samples in the future, in order to study other phase functions (e.g. \cite{heitz2015sggx}) and volume parametrizations. Finally, we plan to integrate RAW imagery, as already done in \cite{Nakath_2021_ICCV}, to linearly relate the datasets to the underlying physical processes.

\section*{Acknowledgement}
This work has received funding from the European Union’s Horizon 2020 research and innovation programme under Grant Agreement No 101000858 (TechOceanS) and the German Research Foundation (Deutsche Forschungsgemeinschaft, DFG) Projektnummer 396311425, through the Emmy Noether Programme.
\FloatBarrier

{
    \small
    \bibliographystyle{ieeenat_fullname}
    \bibliography{main}
}
\clearpage
\setcounter{page}{1}
\maketitlesupplementary
\section{Introduction}
In this supplement to the paper \textit{-- Visual Tomography -- Physically Faithful Volumetric Models of Partially Translucent Objects}, we will present the following:
First, we will take a closer look at the construction of the datasets, especially the markers, the differences between the training and validation sets and the cropping scheme will be displayed in detail.
Then, the idea to initialize a physically-based medium from an emissive one is revisited.
Afterwards, the re-renderings are presented in more detail, followed by a detailed look at the results.
Finally, the last section describes the datasets and videos, which jointly form the supplementary material for this paper.

\section{A closer look at the construction of the datasets}

\begin{figure}[t]
	\begin{center}
		\includegraphics[width=1.\linewidth]{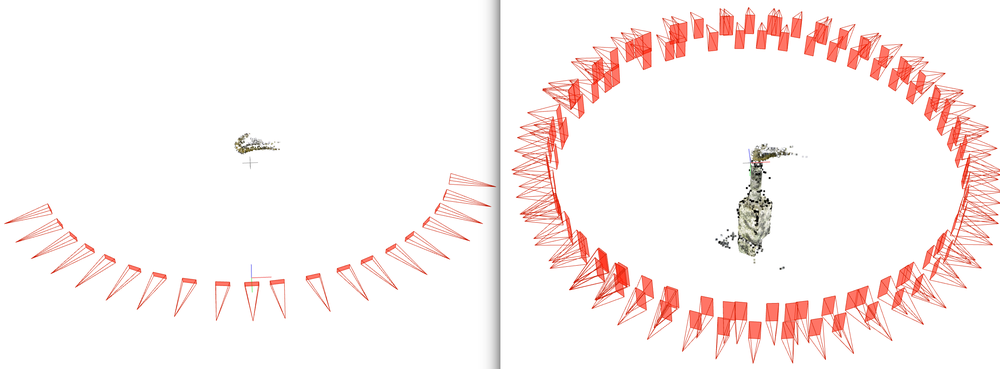} %
	\end{center}
	\caption{Dwarf Prawns, Effect of adding dot markers to the resin base: Left covered dot markers, right, with dot markers. In addition, please note that the camera-poses for training (upper-row), and validation (lower-row) can be observed as well.}
	\label{fig:marker_no_marker_dp}
\end{figure}

\begin{figure}[t]
	\begin{center}
		\includegraphics[width=1.\linewidth]{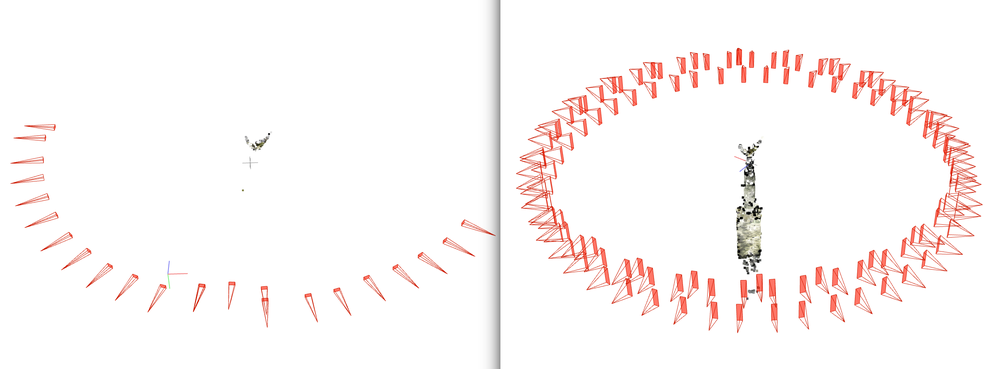}
	\end{center}
	\caption{Praunus Flexuosus: Effect of adding dot markers to the resin base: Left covered dot markers, right, with dot markers. In addition, please note that the camera-poses for training (lower-row), and validation (upper-row) can be observed as well. }
	\label{fig:marker_no_marker_pf}
\end{figure}

\paragraph{The dot markers} \cite{li2013multiple}  are necessary. See the left hand sides of Figs. \ref{fig:marker_no_marker_dp} and \ref{fig:marker_no_marker_pf} for a failed and bended reconstruction of the camera poses, when markers are omitted, while the right hand side shows a complete reconstruction supported by the markers.

\begin{figure}[ht]
	\begin{center}
		\includegraphics[width=.49\linewidth]{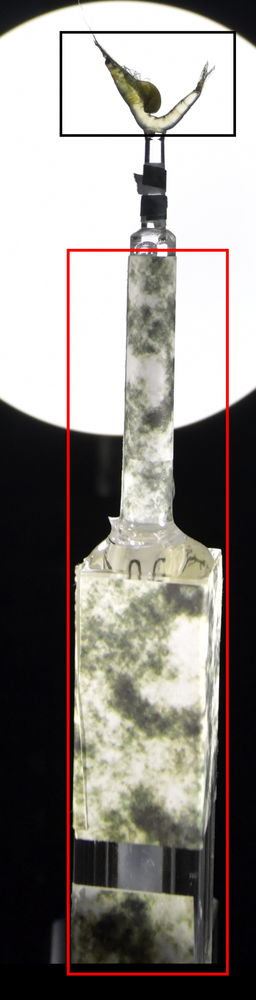}
		\includegraphics[width=.49\linewidth]{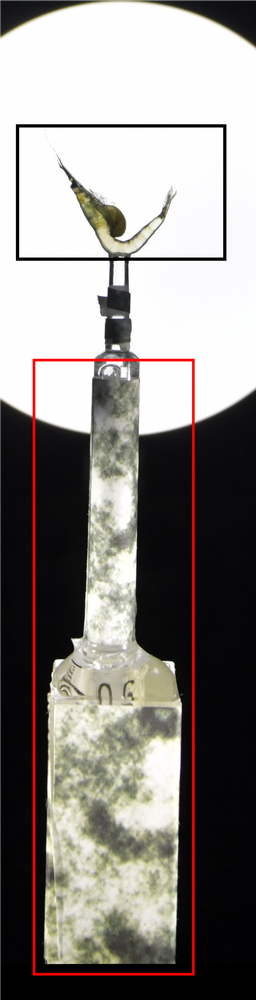}
	\end{center}
	\caption{Praunus Flexuosus: Non-cropped images with cropping area indicated as a black box and dot-markers inside the red box. Left: from the PF-validation set, Right:from the PF-training set.}
	\label{fig:marker_cropping_pf}
\end{figure}

\begin{figure}[ht]
	\begin{center}
		\includegraphics[width=.49\linewidth]{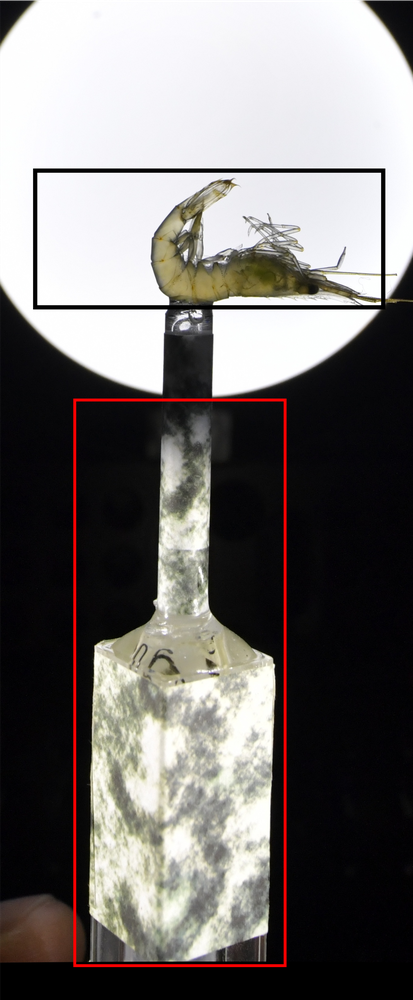}
		\includegraphics[width=.49\linewidth]{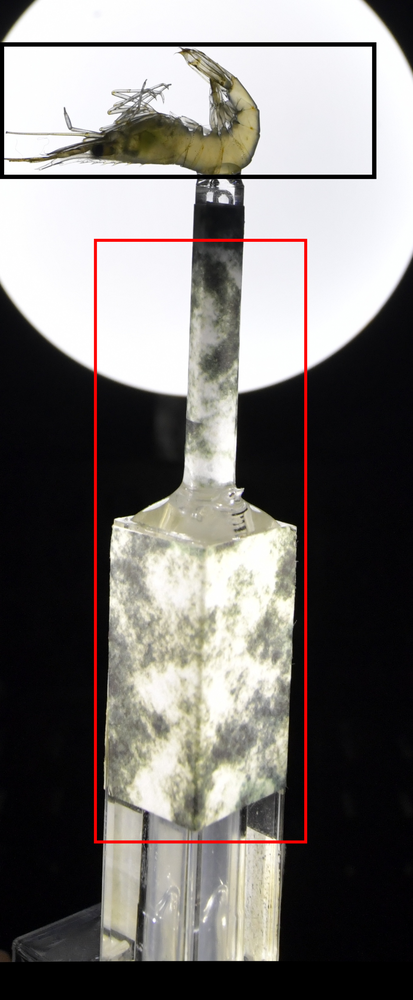}
	\end{center}
	\caption{Dwarf Prawns: Non-cropped images with cropping area indicated as a black box and dot-markers inside the red box. Left: from the DP-validation set, Right:from the DP-training set.}
	\label{fig:marker_cropping_dp}
\end{figure}

\paragraph{A training and a validation set} are taken from the same specimen. See Figs. \ref{fig:marker_no_marker_dp}, \ref{fig:marker_no_marker_pf} for the trajectories and their offsets between the training and validation condition. Furthermore, the respective croppings are drawn in Figs. \ref{fig:marker_cropping_pf}, \ref{fig:marker_cropping_dp}.

\section{Initialization of physically-based medium from a non-physical emissive medium}
\begin{figure*}[t]
	\begin{center}
		\normalfont \small
		\def\svgwidth{1\linewidth}
    \begin{picture}(1,0.20157476)%
      \lineheight{1}%
      \setlength\tabcolsep{0pt}%
      \put(0,0){\includegraphics[width=\unitlength,page=1]{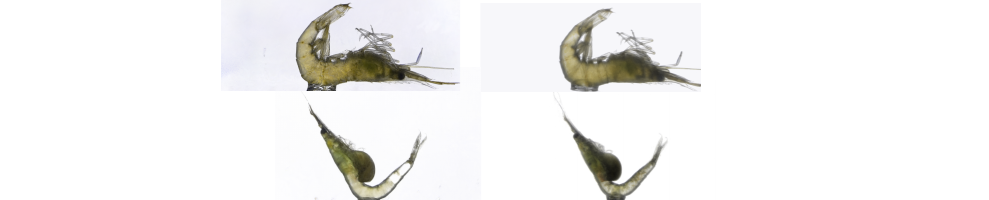}}%
      \put(0.39604241,0.08634537){\color[rgb]{0,0,0}\makebox(0,0)[lt]{\lineheight{1.25}\smash{\begin{tabular}[t]{l}NeRF\end{tabular}}}}%
      \put(0,0){\includegraphics[width=\unitlength,page=2]{inverse_emission.pdf}}%
      \put(0.59499217,0.08812947){\color[rgb]{0,0,0}\makebox(0,0)[lt]{\lineheight{1.25}\smash{\begin{tabular}[t]{l}P-b-volume\end{tabular}}}}%
      \put(0,0){\includegraphics[width=\unitlength,page=3]{inverse_emission.pdf}}%
      \put(0.72492001,0.00506552){\color[rgb]{0,0,0}\makebox(0,0)[lt]{\lineheight{1.25}\smash{\begin{tabular}[t]{l}Physically-based volume\end{tabular}}}}%
      \put(0,0){\includegraphics[width=\unitlength,page=4]{inverse_emission.pdf}}%
      \put(0.00532219,0.00346489){\color[rgb]{0,0,0}\makebox(0,0)[lt]{\lineheight{1.25}\smash{\begin{tabular}[t]{l}Emissive volume\end{tabular}}}}%
      \put(0.00405691,0.04422052){\color[rgb]{0,0,0}\makebox(0,0)[lt]{\lineheight{1.25}\smash{\begin{tabular}[t]{l}Emission $\d c(\theta,\phi)$\end{tabular}}}}%
      \put(0,0){\includegraphics[width=\unitlength,page=5]{inverse_emission.pdf}}%
      \put(0.00475181,0.18263374){\color[rgb]{0,0,0}\makebox(0,0)[lt]{\lineheight{1.25}\smash{\begin{tabular}[t]{l}1D-Density $\sigma$\end{tabular}}}}%
      \put(0,0){\includegraphics[width=\unitlength,page=6]{inverse_emission.pdf}}%
      \put(0.72546161,0.18114829){\color[rgb]{0,0,0}\makebox(0,0)[lt]{\lineheight{1.25}\smash{\begin{tabular}[t]{l}3D-Density $\d \sigma_t$\end{tabular}}}}%
      \put(0,0){\includegraphics[width=\unitlength,page=7]{inverse_emission.pdf}}%
      \put(0.72638787,0.06040503){\color[rgb]{0,0,0}\makebox(0,0)[lt]{\lineheight{1.25}\smash{\begin{tabular}[t]{l}Light radiance $\mathbb{\d L}$\end{tabular}}}}%
      \put(0,0){\includegraphics[width=\unitlength,page=8]{inverse_emission.pdf}}%
      \put(0.43873986,0.00547079){\color[rgb]{0,0,0}\makebox(0,0)[lt]{\lineheight{1.25}\smash{\begin{tabular}[t]{l}Initialize w/ Eq. \ref{eq:inv_em}\end{tabular}}}}%
      \put(0,0){\includegraphics[width=\unitlength,page=9]{inverse_emission.pdf}}%
    \end{picture}%

	\end{center}
	\caption{The idea of initializing densities of the single wavelengths of a physically based volume with the homogeneous density from an emissive volume by using \textit{inverse emission}.}
	\label{fig:inv_em}
\end{figure*}

As shown in Fig. \ref{fig:inv_em}, left, we can locally think of an emissive medium as emitting light over all channels at a certain point in a specific direction $\d c(t, \d d)$, which is subsequently attenuated uniformly over all channels by the respective local densities $\sigma(t)$.
In addition, the density-part of the physically-based medium, which we seek to initialize, is modelled as a local three channel attenuation $\d \sigma_t$.
If we now, e.g., want to yield a red color, we have to emit it in the first medium, while we have to attenuate all colors \textit{but} the red one in the latter medium, i.e., employ \textit{inverse emittance}.
As we previously estimated the light's radiance $\mathbb{\d L}$, we can use this prior knowledge to estimate the attenuation from emittance.
If we additionally take into account a mean local emittance, to average out specularities, we finally yield Eq. \ref{eq:inv_em}
\begin{equation*}
	\d \sigma_t = \sigma(t)  \left(\mathbb{\d L} - \frac{1}{n} \sum\limits_{\theta} \sum\limits_{\phi} \d c(t, \d d)\right).
\end{equation*}
In the center box of Fig. \ref{fig:inv_em}, on the left, the NeRF-optimization result is shown, while the right hand side is the re-rendering of the same pose of the physically-based medium initialized with our mapping.

\section{Re-renderings for different instruments}

\begin{figure*}[ht]
	\begin{center}
		\includegraphics[width=1\linewidth]{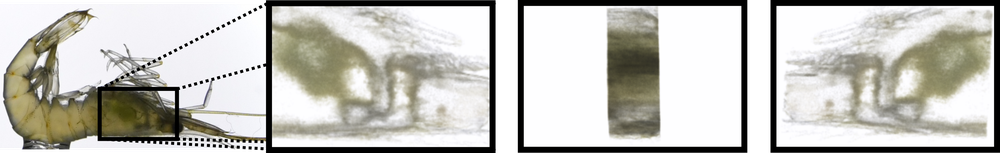}
	\end{center}
	\caption{Dwarf Prawns: Rotating and slicing to get a better glimpse at the intestines of the specimen.}
	\label{fig:slicing}
\end{figure*}

\begin{figure*}[th]
	\begin{center}
		\includegraphics[width=.49\linewidth]{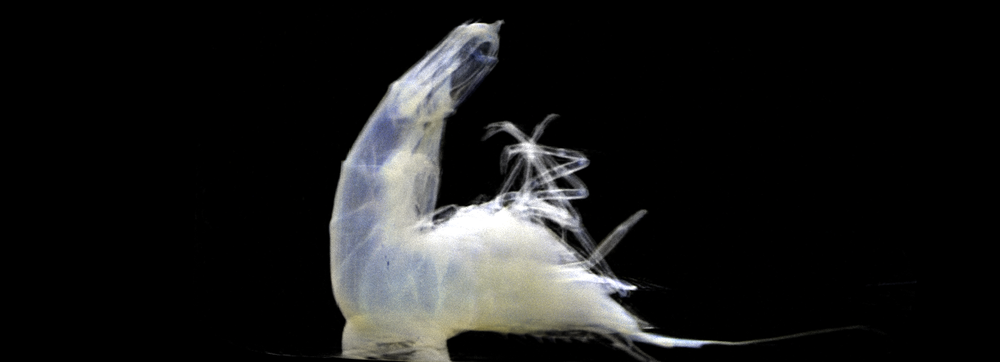}
		\includegraphics[width=.49\linewidth]{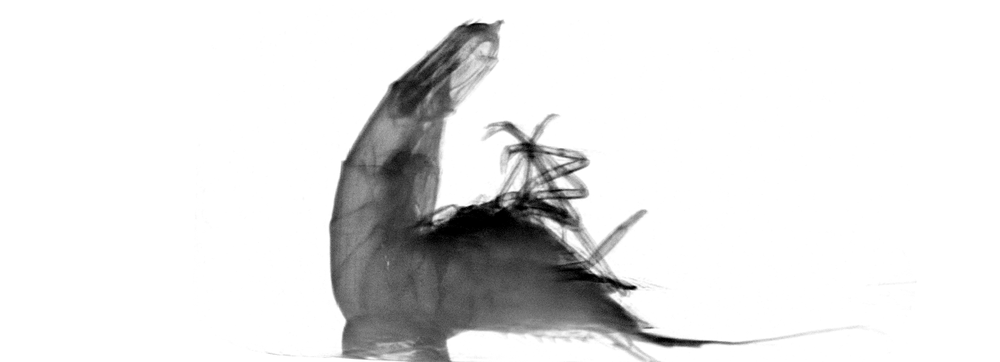}
		\includegraphics[width=.49\linewidth]{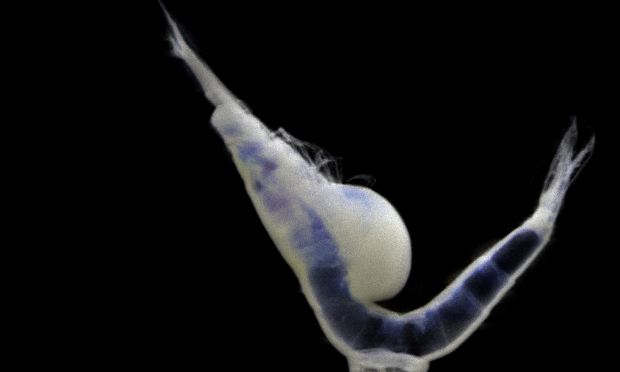}
		\includegraphics[width=.49\linewidth]{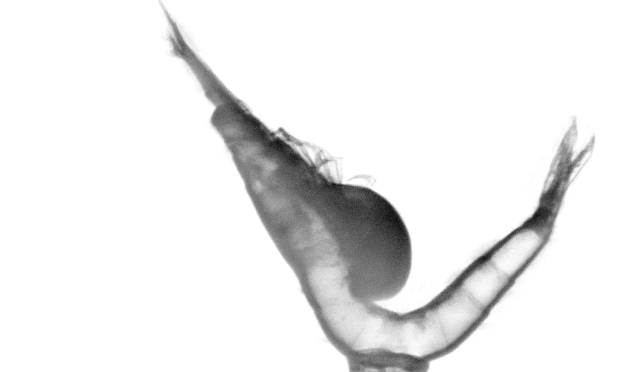}
	\end{center}
	\caption{Synthesized darkfield images and their inverse, as used for the UVP-series plankton sensors. The upper row depicts the Dwarf Prawn, while the lower row shows the Praunus Flexuosus.}
	\label{fig:df}
\end{figure*}

Our pipeline completely disentangles the data from the surrounding, it was recorded in.
As already shown in Fig. \ref{fig:re-renderingpipeline}, we can process the optimized volume to yield improved versatile representations.
In addition, this allows us to e.g., slice and rotate the data to present it in a way, which exposes the intestines, as can be seen in Fig. \ref{fig:slicing}.
Finally,  Fig. \ref{fig:df} shows darkfield images, i.e., light from above, on the left hand side, and the respective inverse darkfield on the right-hand side.
Obviously, the volumes exhibit a very good relighting-capability over the recorded datasets.

\section{A closer look at the results}
In a physically-based scheme it is mandatory to model the scene's parameters as close as possible.
Hence, we model the light sources radiance as optimizable, to account for the actual radiance in combination with the camera's properties (white balance, spectral response, etc. ..).
Not optimizing the light does not only yield a higher residual error due \textit{wrongly colored} background pixels, where only the light source is perceived by the camera: it also results in a worsened volumetric model, as the medium tries to compensate for the specimen as well as the wrongly parametrized background illumination.
In addition, we stop the light optimization early, to avoid an oscillatory behavior of the light's radiance and the parts of the medium trying in turn to  compensate for higher or lower radiances (see Sect. \ref{sec:eval:setup}).
Hence, our optimization over 60 epochs is divided into three phases
\begin{itemize}
	\item Iter 0-10 \textit{Warmup-Phase}: Light and Medium are optimized jointly with a high learning rate of 1e-3,
	\item Iter 10-20 \textit{Main-Phase}: Medium is further build up with a learning rate of 2e-4,
	\item Iter 25-59 \textit{Refinement-Phase}: The medium is refined with a learning rate of 55e-6.
\end{itemize}

\section{Additional Supplementary Material}
Finally, we will provide additional supplementary material, as described in the following:
\begin{itemize}
	\item \textbf{Video:} Convergence over time of the differentiable Renderer for Praunus Flexuosus (/vid/pf\_dr\_lo\_convergence.mp4) and Dwarf Prawns (/vid/dp\_dr\_lo\_convergence.mp4)

	\item \textbf{Video:} Hi-res re-renderings with validation set poses, i.e, novel views, for Praunus Flexuosus (/vid/pf\_dr\_lo\_validation\_hi\_res.mp4) and Dwarf Prawns (/vid/dp\_dr\_lo\_validation\_hi\_res.mp4)

	\item \textbf{Video:} Darkfield re-lighting of the optimized models – i.e., with a light source above the medium for Praunus Flexuosus (/vid/pf\_df.mp4) and Dwarf Prawns (/vid/dp\_df.mp4)

	\item \textbf{Video:} Inverse darkfield as used by the UVP-series plankton sensors for Praunus Flexuosus (/vid/pf\_inv\_df.mp4) and Dwarf Prawns (/vid/dp\_inv\_df.mp4)

	\item \textbf{Video:} Dwarf Prawn immersed in homogeneous medium, i.e., water column, and in darkfield illumination (/vid/dp\_uw\_df.mp4)

	\item \textbf{Video:} Video of sliced re-rendering of intestines as depicted in Fig. \ref{fig:slicing} (/vid/dp\_hi\_res\_re\_renderings\_sliced.mp4)

	\item \textbf{Data:} Full training image sets for Praunus Flexuosus (/data/pf\_train) and Dwarf Prawns(/data/dp\_train), cropped to the actual size used

	\item \textbf{Data:} Full validation image sets for Praunus Flexuosus (/data/pf\_validate) and Dwarf Prawns (/data/dp\_validate), cropped to the actual size used

	\item \textbf{Data:} intrinsic (/data/dp\_cameras.txt)  and extrinsic (/data/dp\_images.txt) camera parameters for  Dwarf Prawns training (Img. IDs 75-122) and validation sets (Img. IDs 25-72) in COLMAP-format

	\item \textbf{Data:} intrinsic (/data/pf\_cameras.txt)  and extrinsic (/data/pf\_images.txt) camera parameters for  Dwarf Prawns training (Img. IDs 25-72) and validation sets (Img. IDs 73-120) in COLMAP-format
\end{itemize}

\end{document}